\documentclass[10pt,twocolumn,twoside]{IEEEtran}
\usepackage{cite}
\usepackage[dvipdfm]{graphicx}
\usepackage[cmex10]{amsmath}
\usepackage{amssymb}
\usepackage{array}
\usepackage{bm}
\usepackage[tight,footnotesize]{subfigure}
\usepackage{stfloats}
\usepackage{inouelab}

\newcommand{\logistic}{\mathop{\mathrm{logistic}}\limits}
\newcommand{\BernoulliDist}{\mathop{\mathrm{Bernoulli}}\limits}
\newcommand{\GammaDist}{\mathop{\mathrm{Gamma}}\limits}
\newcommand{\NormalDist}{\mathop{\calN}\limits}
\newcommand{\stepOO}{{(\!0\!)}}
\newcommand{\steptt}{{(\!t\!)}}
\newcommand{\steptl}{{(\!t\!+\!1\!)}}

\newcommand{\stepinf}{{(\!\infty\!)}}

\begin{document}
\title{Posterior Mean Super-resolution with a Causal Gaussian Markov Random Field Prior}
\author{Takayuki~Katsuki, Akira~Torii, and Masato~Inoue
\thanks{Takayuki~Katsuki, Akira~Torii, and Masato~Inoue are with the Department of Electrical Engineering and Bioscience, Graduate School of Advanced Science and Engineering, Waseda University, 3--4--1, Okubo, Shinjuku, Tokyo 1698555, Japan. E-mail: (see http://www.eb.waseda.ac.jp/m\_inoue/).}%
}

\maketitle

\begin{abstract}
We propose a Bayesian image super-resolution (SR) method with a causal Gaussian Markov random field (MRF) prior. SR is a technique to estimate a spatially high-resolution image from given multiple low-resolution images. An MRF model with the line process supplies a preferable prior for natural images with edges. We improve the existing image transformation model, the compound MRF model, and its hyperparameter prior model. We also derive the optimal estimator -- not the joint maximum a posteriori (MAP) or marginalized maximum likelihood (ML), but the posterior mean (PM) -- from the objective function of the L2-norm (mean square error) -based peak signal-to-noise ratio (PSNR). Point estimates such as MAP and ML are generally not stable in ill-posed high-dimensional problems because of overfitting, while PM is a stable estimator because all the parameters in the model are evaluated as distributions. The estimator is numerically determined by using variational Bayes. Variational Bayes is a widely used method that approximately determines a complicated posterior distribution, but it is generally hard to use because it needs the conjugate prior. We solve this problem with simple Taylor approximations. Experimental results have shown that the proposed method is more accurate or comparable to existing methods.
\end{abstract}


%\IEEEpeerreviewmaketitle

\section{Introduction}
Super-resolution (SR) is an information processing technique that makes it possible to infer a spatially high-resolution (HR) image of a scene from corresponding multiple low-resolution (LR) images that are affected by warping, blurring, and noise. SR can be applied to a variety of images; e.g., still images extracted from several sequential video frames. SR needs the registration of LR images in addition to the image restoration of the registered LR images. Since the earliest work by Tsai and Huang \cite{Tsai1984}, SR has been achieved using various methods \cite{Tipping2003,Babacan2011,Kanemura2007,Kanemura2009,Pickup2007,Hardie1997,Villena2010,Molina2003a,Vandewalle2007} and good overviews of these methods are given in \cite{Borman1998, Farsiu2004, Kang2003, Ng2006,Katsaggelos2007,Milanfar2010}. Generally, SR is an ill-posed inverse problem because inverting the blur process without amplifying the effect of the noise is difficult \cite{Farsiu2004}. In other words, the degrees of freedom of the HR image and pixel-wise observation noise are always higher than the dimensionality of the observed LR images, so complete determination of an HR image is impossible. Therefore, the HR image is frequently inferred as the most preferable image within the framework of the probabilistic information processing, and we handle SR using this framework in this paper. The probabilistic information processing has three key features: 1) model, 2) objective function, and 3) optimization method. In the SR problem, the model includes the observation model and the prior model. The observation model consists of warping, blurring, downsampling, and noise models. The prior model, necessary for the Bayesian framework, mainly consists of an HR image prior, and sometimes includes both the hyperparameter prior for the HR image prior and the registration prior. The objective function evaluates how good or bad an estimator is. The estimator usually represents the inferred HR image, and sometimes includes auxiliary parameters; e.g., the registration parameters and edge information. The optimization method numerically maximizes/minimizes the objective function and determines the estimator. An optimization method is not necessary for simple problems in which an analytical exact solution can be obtained. In the probabilistic information processing, SR can be categorized according to these three key features.

To deal with warping, blurring, and downsampling, a linear transformation model is frequently used \cite{Tipping2003,Kanemura2007,Kanemura2009,Babacan2011}. Warping is usually limited with planar rotation and parallel translation. Blurring is defined by using a point spread function (PSF); a square or Gaussian type PSF is common. Downsampling denotes sampling from an HR image to construct an LR image. Downsampling sometimes includes anti-aliasing. Since these three transformations are linear, they can be combined into a single transformation matrix. As for the noise model, pixel-independent additive white Gaussian noise (AWGN) is usually employed.

The Bayesian framework, especially the HR image prior, is quite useful for SR. The HR image prior provides appropriate smoothness between neighboring pixel luminances. A common type of HR image prior imposes an L2-norm penalty on differences between horizontally and vertically adjacent pixel luminances (the first derivative). The L1-norm of the first derivative is sometimes used, and it has the advantage of robust inference against outliers. The total variation (TV) prior \cite{Babacan2011} employs the L1-norm of the gradient vector. The Huber prior \cite{Pickup2007} is a mixture prior of L1- and L2-norms. The SAR model \cite{Hardie1997,Molina2003b,Villena2010} employs the response of a two-dimensional Laplacian filter (the second derivative). The Gaussian process prior \cite{Tipping2003} has neighboring pixels spread according to a Gaussian distribution. Besides the degree of smoothness between neighboring pixels, information regarding the discontinuity, or equivalently, the edges or line process, is also useful for inference. A common type of prior implementing edges is the compound Markov random field (MRF) prior that was introduced by Geman \& Geman \cite{Geman1984} and is widely used \cite{Molina2003a,Kanemura2007,Kanemura2009}. With respect to the compound MRF \cite{Chellappa1993,Jeng1991} prior, the normalizing constant, or equivalently, the partition function, is usually difficult to calculate because it has an exponential calculation cost with respect to the dimensionality of the line process. Recently, Kanemura \textit{et al.} \cite{Kanemura2007,Kanemura2009} confusingly introduced a ``causal'' type of Gaussian MRF prior whose calculation cost is polynomial. We try to improve this prior in this paper.

The SR estimator should be derived from an objective function. As the objective function, a posterior distribution has been widely employed. Since the posterior distribution usually includes both the HR image and registration parameters, the joint maximum a posteriori (MAP) solution \cite{Hardie1997} is a suitable estimator for this objective function. Other than the joint MAP, the use of the marginalized maximum likelihood (ML) \cite{Tipping2003,Kanemura2007} or marginalized MAP \cite{Pickup2007} has been proposed. Tipping \textit{et al.} \cite{Tipping2003} and Kanemura \textit{et al.} \cite{Kanemura2007,Kanemura2009} determine the registration parameters by using ML inference, where the HR image is marginalized out, and determine the HR image by using MAP inference. Pickup \textit{et al.} \cite{Pickup2007} determines the HR image by using MAP inference, wherein the registration uncertainties are marginalized out, and assumes that the registration parameters are pre-registered by using standard registration techniques. Marginalized ML is also called type-II ML, evidence approximation, or empirical Bayes. Marginalized ML has no registration prior, unlike marginalized MAP. Pickup \textit{et al.} \cite{Pickup2007} reported that marginalized MAP is superior to both joint MAP and marginalized ML. We evaluate the accuracy of SR methods in terms of the L2-norm (mean square error) -based peak signal-to-noise ratio (PSNR). Therefore, we think it is natural to employ PSNR as the objective function. For this objective function, posterior mean (PM) is a suitable estimator. The variational Bayes \cite{Attias1999} approach \cite{Babacan2011} seems to approximately determine the PM of the HR image, although the authors assume some registration parameters are known and use point-estimate model parameters obtained by ML inference. To determine the exact PM of the HR image, all parameters other than the HR image should be marginalized out over the joint posterior distribution.

The type of optimization method to use is not as substantial a problem as the choice of model and objective function, but it is still important. Since almost all good estimators cannot be exactly determined because of difficult analytical integration or an exponential calculation cost, some approximation methods need to be introduced. Also, parameter tuning is necessary in many numerical optimization methods; e.g., of the initial value and the step-width settings in gradient methods. Specifically, in early work done on image restoration, an annealing method was used for the joint MAP solution \cite{Geman1984,Jeng1990}. For marginalized ML and marginalized MAP solutions, the scaled conjugate gradients algorithm was used \cite{Tipping2003,Pickup2007}. In recent work, the variational expectation-maximization (EM) algorithm has been applied, which includes the gradient method in the M step \cite{Kanemura2007,Kanemura2009}. The variational Bayes approach has also been applied \cite{Babacan2011}. This method includes nested optimization of the majorization-minimization approach. This majorization-minimization approach seems to affect both the HR image prior and the estimator. Specifically, it modifies the TV prior to include a discontinuity parameter (called local spatial activity). In addition, this parameter is point-estimated when the HR image is inferred.

In this paper, we propose a new SR method that employs a ``causal'' Gaussian MRF prior and utilizes variational Bayes to calculate the optimal estimator, PM, with respect to the objective function of the L2-norm-based PSNR. This is a straightforward approach, but it was not proposed earlier possibly because an important limitation of variational Bayes is that a conjugate prior is needed. We solve this problem through simple Taylor approximations. In Section II, we define models, where we introduce a novel unified warping, blurring and downsampling model, an improved HR image prior, an improved hyperparameter prior, and a registration prior. In Section III, we employ PSNR as the objective function and derive the optimal estimator, PM, from this objective function. In Section IV, we determine the PM by using variational Bayes and Taylor approximations. In Section V, we evaluate the proposed method by comparing it with existing methods. We discuss the proposed method in Section VI and conclude in Section VII.

\section{Model}
\subsection{Definitions}
First, we define the gamma, Bernoulli, and Gaussian distributions used in this paper:
\begin{align}
	\GammaDist(x; a, b)
	&\equiv \frac{b^a}{\Gamma(a)} x^{a-1} \rme^{-b x} \quad (x > 0), \nonumber\\
	\BernoulliDist(x; \mu)
	&\equiv \mu^x (1-\mu)^{1-x} \quad (x \in \{ 0, 1 \}), \nonumber\\
	\NormalDist(\bx; \bmu, \bSigma)
	&\equiv |2\rmpi\bSigma|^{-\frac{1}{2}} \rme^{-\frac{1}{2} (\bx-\bmu)^\top \bSigma^{-1} (\bx-\bmu)} \quad (\bx \in \mathbb{R}^d), \nonumber
\end{align}
Here, $\Gamma$ is the gamma function, $|\bullet|$ denotes the determinant of a given matrix, superscript $\top$ denotes the transpose, $\mathbb{R}$ is the real number field, and $d$ is the dimension of $\bx$. The logistic function and Kullback-Leibler (KL) divergence from distributions $p(\bx)$ to $q(\bx)$ are respectively defined as
\begin{align}
	\logistic(x)
	&\equiv \frac{1}{1 + \rme^{-x}}, \nonumber\\
	D_\mathrm{KL}(p(\bx) \| q(\bx))
	&\equiv \left\langle \ln\frac{p(\bx)}{q(\bx)} \right\rangle_{p(\bx)}, \nonumber
\end{align}
where the angle brackets $\langle\bullet\rangle_\circ$ denote the expectation of $\bullet$ with respect to a distribution $\circ$. Additionally, $\tr$ denotes the trace of a given matrix. $\mathrm{diag}$ denotes a diagonal matrix. $\bI$ is an identity matrix of appropriate size. $\bm{0}$ is a zero vector or a zero matrix of appropriate size. All the vectors in this paper are column vectors. The $\|\bullet\|_2$ denotes the L2-norm of a given vector. At this point, these variables have absolutely nothing to do with the variables that appear later.

\subsection{Observation Model}

\begin{figure}[tb]
	\centering
	\includegraphics[width=8cm]{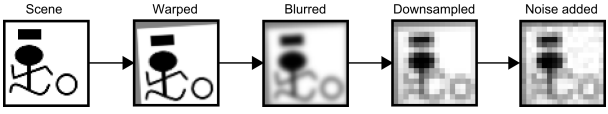}
	\caption{An illustration of the image observation process}
	\label{FigProcess}
\end{figure}

Our task is to estimate an HR grayscale image, $\bx \in \mathbb{R}^{N_\bx}$, from the observed multiple LR grayscale images, $\bY \equiv \{ \by_l \}_{l=1}^L, \by_l \in \mathbb{R}^{N_\by}$. Images $\by_l$ and $\bx$ are regarded as lexicographically stacked vectors. The number of pixels for each LR image, $N_\by$, is assumed to be less than that of the HR image, $N_\bx$; i.e., $N_\by < N_\bx$. We do this estimation using an SR technique whose resolution enhancement factor is $\alpha \equiv \sqrt[]{N_\bx / N_\by}$ $(>1)$. Although we define the range of a pixel luminance value as infinite, we use $-1$ for black, $+1$ for white, and values between $-1$ and $+1$ for gradual gray.

The image observation process is modeled as shown in Fig. \ref{FigProcess}; the HR image $\bx$ is geometrically warped, blurred, downsampled, and corrupted by noise $\bepsilon_l$ to form the observed LR image $\by_l$:
\begin{align}
	\label{EqImageObservationProcess}
	\by_l \equiv \bW(\bphi_l)\bx + \bepsilon_l,
\end{align}
or, more strictly,
\begin{align}
	\label{EqObservationModel}
	p(\bY | \bx, \beta, \bPhi)
	\equiv \prod_{l=1}^L \NormalDist(\by_l; \bW(\bphi_l) \bx, \beta^{-1}\bI).
\end{align}
The $\bepsilon_l \in \mathbb{R}^{N_\by}$ is AWGN with precision (inverse variance) $\beta$ $(>0)$. Here, $\bW(\bphi_l)$ is the $N_\by \times N_\bx$ transformation matrix that is simultaneously used for warping, blurring, and downsampling. It is defined as
\begin{align}
	\label{EqTransformationMatrix}
	\bW(\bphi_l)_{j,i}
	&\equiv \frac{                   \NormalDist\Big(\vec{\chi}\big(\theta_l,\vec{o}_l,\vec{\zeta}_j,\vec{\xi}_i\big); 0, \gamma_l^{-1} \bI\Big)}
	             {\sum_{i^\prime \in \calI} \NormalDist\Big(\vec{\chi}\big(\theta_l,\vec{o}_l,\vec{\zeta}_j,\vec{\xi}_{i^\prime}\big); 0, \gamma_l^{-1} \bI\Big)}, \\
	\vec{\chi}\big(\theta,\vec{o},\vec{\zeta},\vec{\xi}\big)
	&\equiv 
	\begin{bmatrix}
		 \cos \theta & \sin \theta \\
		-\sin \theta & \cos \theta
	\end{bmatrix}
	\left( \alpha \vec{\zeta} - \vec{o} \right) - \vec{\xi},
\end{align}
where $\calI$ represents the extent of the summation (explained in the next paragraph), and the vectors $\vec{\xi}_i$ and $\vec{\zeta}_j$ respectively denote the two-dimensional positions of the $i$-th pixel of the original HR image and the $j$-th pixel of the observed LR image. We define the center of each image as the origin and the size of each pixel is $1$ by $1$. For example, regarding an HR image with $40 \times 40$ pixels, each $\vec{\xi}$ represents $[-19.5,-19.5]^\top, [-18.5,-19.5]^\top, ...,[19.5,19.5]^\top$. $\theta_l$ and $\vec{o}_l$ represent the warping parameters of the $l$-th LR image: the rotational motion parameter and translational motion parameter. The Gaussian distribution in (\ref{EqTransformationMatrix}) represents a Gaussian PSF that defines the blur, and $\gamma_l$ $(>0)$ represents its precision parameter. In this paper, we assume $\gamma_l$ also differs for each observed image. These transformation parameters are packed into $\bphi_l$, which is defined as
\begin{align}
	\bPhi \equiv \{ \bphi_l \}_{l=1}^L, ~
	\bphi_l
	\equiv [ \phi_{l,k} ]_{k=1}^4
	\equiv [ \theta_l, [\vec{o}_l]_h, [\vec{o}_l]_v, \gamma_l ]^\top,
\end{align}
where subscripts $h$ and $v$, respectively, denote horizontal and vertical positions on the image.

In previous works \cite{Tipping2003,Kanemura2007,Kanemura2009}, the extent of $\calI$ was defined as the extent of the HR image. According to this definition, however, the shape of the PSF is no longer Gaussian. For example, at the corner of the HR image, the shape is not omnidirectional but limited in a way such as that of a quadrant. In this paper, the extent of $\calI$ is defined as infinite, and the luminance values outside the HR image are defined as $0$ (middle gray). This normalization term faithfully represents the Gaussian PSF. We also found that this normalization term is exactly given by using the elliptic theta function $\vartheta_3$, and we can rewrite $\bW(\bphi_l)$ as
\begin{align}
	\label{EqTransformationMatrixEllipticTheta}
	&\bW(\bphi_l)_{j,i}\nonumber\\
	&= \!\frac{\NormalDist\Big(\vec{\chi}\big(\theta_l,\vec{o}_l,\vec{\zeta}_j,\vec{\xi}_i\big); 0, \gamma_l^{-1} \bI\Big)}
		{\vartheta_3\! \left(\! \Big[ \vec{\chi}\!\big(\!\theta_l\!,\vec{o}_l\!,\vec{\zeta}_j\!,\vec{\xi}_i\!\big)\! \Big]_h, \!\rme^{-\frac{2\rmpi^2}{\gamma_l}} \!\right)
		 \!\vartheta_3\! \left(\! \Big[ \vec{\chi}\!\big(\!\theta_l\!,\vec{o}_l\!,\vec{\zeta}_j\!,\vec{\xi}_i\!\big)\! \Big]_v,\! \rme^{-\frac{2\rmpi^2}{\gamma_l}} \!\right)}, \\
	&\vartheta_3(u, q) \equiv 1 + 2 \sum_{n=1}^\infty q^{n^2} \cos 2 n \rmpi u.
\end{align}
The elliptic theta function includes an infinite series, but it is easily determined numerically because the convergence is quite fast. In (\ref{EqTransformationMatrixEllipticTheta}), the normalization term (the denominator of the right-hand side) seems to depend on $i$ because $ \vec{\chi}\big(\theta_l,\vec{o}_l,\vec{\zeta}_j,\vec{\xi}_i\big)$ includes $\vec{\xi}_i$, but this is not true. Because the elliptic theta function is a periodic function with respect to the argument $u$ with period $1$, and $ \vec{\chi}\big(\theta_l,\vec{o}_l,\vec{\zeta}_j,\vec{\xi}_i\big)$ can only take discrete values with step size $1$ for the horizontal and vertical directions, the normalization term has the same value with respect to $i$.

\subsection{HR Image Prior}
Here, we introduce a ``causal'' Gaussian MRF prior for the HR image and additional latent variables. These latent variables are called the line process that controls the local correlation among pixel luminances. The introduction of the latent variables enables explicit expression of the possible discontinuity in the HR image. The line process, $\bmeta$, consists of binary variables $\eta_{i,j} \in \{ 0,1 \}$ for all adjacent pixel pairs $i$ and $j$. Its size equals $N_\bmeta \equiv 2 N_\bx - [\textrm{number of HR image's horizontal pixels}] - [\textrm{number of HR image's vertical pixels}]$. We define the prior as
\begin{align}
	\label{EqPriorDistributionXEta}
	&p(\bx, \bmeta | \lambda, \rho, \kappa)
	\equiv p(\bx | \bmeta, \rho, \kappa) p(\bmeta | \lambda) \\
	&= \exp \Bigg[ -\lambda \sum_{i \sim j} (1 \!-\! \eta_{i,j}) - \frac{\rho}{2} \sum_{i\sim j}\eta_{i,j}(x_i\!-\!x_j)^2 - \frac{\kappa}{2} \|\bx\|_2^2 \nonumber\\
	&~~~~~~~~~~~ +\frac{1}{2} \ln \left| \frac{\bA(\bmeta, \rho, \kappa)}{2\rmpi} \right| + N_\bmeta \ln \logistic(\lambda) \Bigg],
\end{align}
where
\begin{align}
	\label{EqPriorDistributionEta}
	p(\bmeta | \lambda)
	&\equiv \prod_{i \sim j} \BernoulliDist \left( \eta_{i,j}; \logistic(\lambda) \right), \\
	p(\bx | \bmeta, \rho, \kappa)
	&\equiv \NormalDist(\bx; \bm{0}, \bA(\bmeta, \rho, \kappa)^{-1}), \\
	\label{EqAMatrix}
	\bA(\bmeta, \rho, \kappa)_{i,j}
	&\equiv
	\begin{cases}
		\rho \sum_{k \sim i} \eta_{i,k} + \kappa, & i=j, \\
		-\rho \eta_{i,j},                         & i \sim j, \\
		0,                                        & \mathrm{otherwise}.
	\end{cases}
\end{align}
Here, the summation $\sum_{i \sim j}$ is taken over all pairs of adjacent pixels. The notation $i \sim j$ means that the $i$-th and $j$-th pixels are adjacent in the upward, downward, leftward, and rightward directions. The line process $\bmeta$ switches the local characteristics of the prior. It indicates whether two adjacent pixels take similar values or independent values. When $\eta_{i,j} = 1$, the $i$-th and the $j$-th pixels are strongly smoothed according to the quadratic penalty, whereas there is no smoothing when $\eta_{i,j} = 0$. The hyperparameter $\lambda$ $(>0)$ is an edge penalty parameter that prevents $\eta_{i,j}$ from excessively taking edges. Note that $\lambda$ is restricted to positive values because a negative $\lambda$ leads to a reward rather than a penalty for taking edges. $\rho$ $(>0)$ is a smoothness parameter that prevents the differences in adjacent pixel luminances from becoming large, and $\kappa$ $(>0)$ is a contrast parameter that prevents $\bx$ from taking an improperly large absolute value. On the other hand, in previous works \cite{Kanemura2007,Kanemura2009}, $\kappa$ is assumed to be $0$, which results in an improper normalizing constant (see Discussion). $\bA(\bmeta, \rho, \kappa)$ is the $N_\bx \times N_\bx$ precision matrix of $\bx$.

We have defined the introduced causal Gaussian MRF prior in the joint distribution form of $\bx$ and $\bmeta$, i.e., $p(\bmeta)p(\bx|\bmeta)$. We call such a model ``causal'' because $\bmeta$ seems to cause $\bx$. The MRF model is defined as having the property
\begin{align}
	\label{EqMRFX}
	p(x_i | \bx \backslash x_i, \bmeta) = p(x_i | \bx_{\calL(i)}, \bmeta_{i,\calL(i)})
\end{align}
in this case; i.e., the conditional distribution of a random variable, $x_i$, given all other variables, $\bx \backslash x_i$ and $\bmeta$, equals the conditional distribution of the random variable, $x_i$, given its ``neighboring'' variables, $\bx_{\calL(i)}$ and $\bmeta_{i,\calL(i)}$. If this conditional distribution is a Gaussian distribution, such an MRF is called a Gaussian MRF.

The ``compound'' MRF prior is usually defined in the form of the Gibbs distribution \cite{Geman1984},
\begin{align}
	\label{EqPriorDistributionAnotherXEta}
	&\tp(\bx, \bmeta)
	\equiv \frac{ \exp(-\tH(\bx, \bmeta)) }{ \sum_\bmeta \int \exp(-\tH(\bx, \bmeta)) \mathrm{d}\bx },
\end{align}
which is based on some microstate energy function, or equivalently, a Hamiltonian, such as
\begin{align}
	\label{EqEnergyFunction}
	&\tH(\bx, \bmeta) \nonumber\\
	&\equiv \lambda \sum_{i \sim j} (1 \!-\! \eta_{i,j}) + \frac{\rho}{2} \sum_{i\sim j}\eta_{i,j}(x_i\!-\!x_j)^2 + \frac{\kappa}{2} \|\bx\|_2^2.
\end{align}
In addition to the property of (\ref{EqMRFX}), a compound MRF also has the property of
\begin{align}
	\label{EqMRFEta}
	\tp(\eta_{i,j} | \bx, \bmeta \backslash \eta_{i,j}) = \tp(\eta_{i,j} | x_i, x_j),
\end{align}
whereas the introduced ``causal'' Gaussian MRF prior does not. Therefore, we do not call the introduced prior a ``compound'' MRF prior, even though (\ref{EqPriorDistributionXEta}) and (\ref{EqPriorDistributionAnotherXEta}) have similar forms. Furthermore, the introduced ``causal'' Gaussian MRF prior is a generative model, whereas the ``compound'' MRF is not. A generative model has the advantage of reducing the calculation cost (see Discussion).

\subsection{Hyperparameter Prior}
Generally, prior distributions should be non-informative unless we have explicit reasons because an informative prior leads to heuristics. Actually, we define the prior distributions for the hyperparameters of the HR image prior to be as non-informative as possible: 
\begin{align}
	\label{EqPriorDistributionLambda}
	&p(\lambda, \rho, \kappa, \beta) \equiv
	\GammaDist(\lambda; a_\lambda^\stepOO, b_\lambda^\stepOO)
	\GammaDist(\rho; a_\rho^\stepOO, b_\rho^\stepOO) \nonumber\\
	&\hphantom{p(\lambda, \rho, \kappa, \beta) \equiv}\times
	\GammaDist(\kappa; a_\kappa^\stepOO, b_\kappa^\stepOO)
	\GammaDist(\beta; a_\beta^\stepOO, b_\beta^\stepOO), \\
	&a_\lambda^\stepOO \equiv 10^{-2}, b_\lambda^\stepOO \equiv 10^{-2}, ~
	 a_\rho   ^\stepOO \equiv 10^{-2}, b_\rho   ^\stepOO \equiv 10^{-2}, \nonumber\\
	&a_\kappa ^\stepOO \equiv 10^{-2}, b_\kappa ^\stepOO \equiv 10^{-2}, ~
	 a_\beta  ^\stepOO \equiv 10^{-2}, b_\beta  ^\stepOO \equiv 10^{-2}.
\end{align}
For a gamma distribution, the number of effective prior observations in the Bayesian framework is equal to two times parameter $a$. As shown in the Appendix, the number of observations for the hyperparameter $\lambda$ is $N_\bmeta$ in this SR. Also, that for $\rho$ and $\kappa$ is $N_\bx$, and that for $\beta$ is $L N_\by$. Therefore, the above settings -- e.g., $2a_\lambda^\stepOO \ll N_\bmeta$ -- are considered sufficiently non-informative.
Superscript $\stepOO$ is added because we use these parameters as the initial values of variational Bayes later.

\subsection{Registration Prior}
For the registration parameters including the blurring parameter, we also define the corresponding prior as
\begin{align}
	\label{EqPriorDistributionPhi}
	&p(\bPhi) \equiv \prod_{l=1}^L \NormalDist(\bphi_l; \bmu_{\bphi_l}^\stepOO, \bSigma_{\bphi_l}^\stepOO), \\
	&\bmu_{\bphi_l}^\stepOO \equiv [0, 0, 0, 12/\alpha^2], ~
	 \bSigma_{\bphi_l}^\stepOO \equiv \mathrm{diag}[10^{-3}, 10^{0}, 10^{0}, 10^{-3}].
\end{align}
For the rotational motion parameter $\theta_l$, the prior assumes $0 \pm 1.81$ degree ($\frac{180}{\rmpi} \sqrt{10^{-3}} \simeq 1.81$). This assumption is considered suitable for this SR task. Similarly, an assumption of $0 \pm 1$ pixels for translational motion parameters $[\vec{o}_l]_h$ and $[\vec{o}_l]_v$ is considered suitable. For blurring parameter $\gamma_l$, $\mu_{\gamma_l}^\stepOO$ is taken to be the value equivalent to the anti-aliasing of the scale factor $\alpha$.

\section{Objective Function and Estimator}
\subsection{Peak Signal-to-Noise Ratio (PSNR)}
First, we confirm that the joint distribution of all random variables can now be explicitly given as 
\begin{align}
	p(\bY, \bz)
	&= p(\bY | \bx, \beta, \bPhi) p(\bx, \bmeta | \lambda, \rho, \kappa) p(\lambda, \rho, \kappa, \beta) p(\bPhi), \\
	\bz
	&\equiv [ \bx, \bmeta, [\lambda, \rho, \kappa, \beta], \bPhi ],
\end{align}
Once the joint distribution is obtained, we can derive all the marginal and conditional distributions; e.g., the posterior distribution $p(\bz | \bY)$ and joint distribution of the HR and LR images $p(\bY, \bx)$.

One of the most commonly used evaluation functions of the inferred image is the L2-norm (mean square error) -based PSNR. It is defined as
\begin{align}
	\mathrm{PSNR}(\hbx; \bx) \equiv 10 \log_{10} \frac{2^2}{\frac{1}{N_\bx} \| \hbx - \bx \|_2^2},
\end{align}
where $\hbx$ is the estimator of the HR image and $\bx$ is the true HR image. Since only LR images, $\bY$, are available for the estimator, we sometimes explicitly express it as a function form, $\hbx(\bY)$. Now, our objective function (functional) to be maximized regarding the estimator is defined as
\begin{align}
	F(\hat{\bm{x}}(\bm{Y}))\equiv 10 \log_{10} \frac{2^2}{\left\langle\frac{1}{N_{\bm{x}}} \| \hat{\bm{x}}(\bm{Y}) - \bm{x} \|_2^2\right\rangle_{p(\bm{Y},\bm{x})}}.
\end{align}
This is because we prefer good estimator performance on average over various HR images and the corresponding LR images. Here, we assume that the occurrence rate of HR and LR images exactly coincides with the model we just introduced.

\subsection{Posterior Mean (PM)}
Using the above objective function, we can explicitly derive the best estimator of the HR image as the PM,
\begin{align}
	\argmax_{\hat{\bm{x}}(\bm{Y})} F(\hat{\bm{x}}(\bm{Y}))
	= \left\langle \bm{x} \right\rangle_{p(\bm{x} | \bm{Y})}.
\end{align}
Here, we used the well-known fact that the PM coincides with the minimum mean square error estimator in Bayesian framework. Note that $p(\bx | \bY)$ needs marginalization of all parameters other than $\bx$ over $p(\bz | \bY)$. If the PM of the line process or other model parameters is necessary, it can also be determined in the same manner.

\section{Optimization Method}
\subsection{Variational Bayes}
Though we could derive the optimal estimator, we cannot obtain the analytical solutions of the posterior distribution $p(\bz | \bY)$ and marginalized posterior distribution $p(\bx | \bY)$. Consequently, we have to rely on approximations. Here, we employ variational Bayes.

Variational Bayes \cite{Attias1999} provides a trial distribution $q(\bz)$ that approximates the true posterior. We impose a factorization assumption on the trial distribution,
\begin{align}
	q(\bz)
	\equiv q(\bx) q(\bmeta) q(\lambda, \rho, \kappa, \beta) q(\bPhi).
\end{align}
Note that, at this moment, the distribution family of each factorized distribution is not limited. We identify the optimal trial distribution that minimizes the KL divergence between the trial and the true distributions as the best approximation of the true distribution:
\begin{align}
	\hq(\bz)
	\equiv \argmin_{q(\bz)} D_\mathrm{KL}(q(\bz) \| p(\bz | \bY)).
\end{align}
Actually, the trial distribution that minimizes the KL divergence, not from $q(\bz)$ to $p(\bz | \bY)$ but from $p(\bz | \bY)$ to $q(\bz)$ coincides with the product of the exact marginal distributions as
\begin{align}
	\argmin_{q(\bz)} D_\mathrm{KL}(p(\bz | \bY) \| q(\bz))
	= \prod_i p(z_i | \bY),
\end{align}
but this minimization is difficult to calculate.

Under the factorization assumption of the trial distribution and the extremal condition of the KL divergence, each optimal trial distribution should satisfy the self-consistent equations,
\begin{align}
	\label{EqSelfConsistent}
	\hq(z_i)
	&\propto \exp \langle \ln p(\bz | \bY) \rangle_{ \prod_{j \neq i} \hq(z_j)}.
\end{align}
In the common style of variational Bayes \cite{Bishop2003,Babacan2011}, this equation is solved by making repetitive updates,
\begin{align}
	\label{EqRecurrenceInitiation}
	q^\stepOO(z_i)
	&\equiv p(z_i), \\
	\label{EqRecurrenceRelation}
	q^\steptl(z_i)
	&\propto \exp \langle \ln p(\bz | \bY) \rangle_{ \prod_{j \neq i} q^\steptt(z_j)}.
\end{align}
Each factorized trial distribution is supposed to converge to the optimal distribution. Sometimes, some $q^\steptl(z_j)$s are used instead of $q^\steptt(z_j)$s for the distribution on the right-hand side of (\ref{EqRecurrenceRelation}). It depends on the hierarchical structure of the model. Similarly, some $q^\stepOO(z_i)$s may not be necessary.

\subsection{Taylor Approximations}
Although variational Bayes is a widely used general framework, its application is difficult in practice because it requires a conjugate prior. The prior distributions we have introduced are not conjugate priors. However, we have found that simple Taylor approximations make them conjugate and enable the analytical exact expectations in (\ref{EqRecurrenceRelation}).

Here, to simplify the notation, we define the mean values of the latent variables $\bmeta$, the hyper parameters $\lambda, \rho, \kappa,\beta$, and the registration parameters $\bphi_l$ over the trial distributions in the step number $t$ of the updates of variational Bayes as $\bmu_{\bmeta}^\steptt$, $\mu_{\lambda}^\steptt$, $\mu_{\rho}^\steptt$, $\mu_{\kappa}^\steptt$, $\mu_{\beta}^\steptt$, $\bmu_{\bphi_l}^\steptt$.

Specifically, we use first-order Taylor approximations for three non-linear terms. $\bW(\bphi_l)$ is approximated around $\bphi_l = \bmu_{\bphi_l}^\steptt$,
\begin{align}
	\label{EqTaylorApproximation1}
	&\bW(\bphi_l)
	\simeq \bW^\steptt_l + \sum_{k=1}^4 [\bphi_l - \bmu_{\bphi_l}^\steptt]_k \bW'^\steptt_{l,k},
\end{align}
where
\begin{align}
	\bW^\steptt_l
	&\equiv \bW(\bmu_{\bphi_l}^\steptt), \\
	\bW'^\steptt_{l,k}
	&\equiv \left. \frac{\partial\bW(\bphi_l)}{\partial\phi_{l,k}} \right|_{\bphi_l = \bmu_{\bphi_l}^\steptt}.
\end{align}
Similarly, $\ln\left|\bA(\bmeta, \rho, \kappa)\right|$ is approximated around $[\bmeta, \ln \rho, \ln \kappa] = [\bmu_\bmeta^\steptt, \ln \mu_\rho^\steptt, \ln \mu_\kappa^\steptt]$,
\begin{align}
	\label{EqTaylorApproximation2}
	&\ln \left| \bA(\bmeta, \rho, \kappa) \right|
	\simeq \ln \left| \bA(\bmu_\bmeta^\steptt, \mu_\rho^\steptt, \mu_\kappa^\steptt) \right| \nonumber\\
	&~~ + \tr \bA(\bmu_\bmeta^\steptt, \mu_\rho^\steptt, \mu_\kappa^\steptt)^{-1}
	 \Big[ \mu_\rho^\steptt \bA( \bmeta - \bmu_\bmeta^\steptt, 1, 0) \nonumber\\
	&~~ + ( \ln \rho - \ln \mu_\rho^\steptt ) \mu_\rho^\steptt \bA( \bmu_\bmeta^\steptt, 1, 0)
	    + ( \ln \kappa - \ln \mu_\kappa^\steptt ) \mu_\kappa^\steptt \bI \Big].
\end{align}
We also use a similar approximation around $[\bmeta, \ln \rho, \ln \kappa] = [\bmu_\bmeta^\steptl, \ln \mu_\rho^\steptt, \ln \mu_\kappa^\steptt]$. In addition, $\ln \logistic(\lambda)$ is approximated around $\ln \lambda = \ln \mu_\lambda^\steptt$,
\begin{align}
	\label{EqTaylorApproximation3}
	& \ln \logistic(\lambda)
	 \simeq \ln \logistic(\mu_\lambda^\steptt) \nonumber\\
	&~~ + ( \ln \lambda - \ln \mu_\lambda^\steptt ) \mu_\lambda^\steptt \logistic(-\mu_\lambda^\steptt).
\end{align}

\subsection{Update Equations}
The trial distributions are obtained from (\ref{EqRecurrenceInitiation})-(\ref{EqTaylorApproximation1}), (\ref{EqTaylorApproximation2}), and (\ref{EqTaylorApproximation3}), as follows:
\begin{align}
	\label{EqTrialDistributionEta}
	q^\steptt(\bmeta)
	&= \prod_{i \sim j} \BernoulliDist(\eta_{i,j}; \mu_{\eta_{i,j}}^\steptt), \\
	\label{EqTrialDistributionX}
	q^\steptt(\bx)
	&= \NormalDist(\bx; \bmu_\bx^\steptt, \bSigma_\bx^\steptt), \\
	\label{EqTrialDistributionLambda}
	q^\steptt(\lambda, \rho, \kappa, \beta)
	&= \GammaDist(\lambda; a_\lambda^\steptt, b_\lambda^\steptt)
	   \GammaDist(\rho;    a_\rho   ^\steptt, b_\rho   ^\steptt) \nonumber\\
	&~~~\times
	   \GammaDist(\kappa;  a_\kappa ^\steptt, b_\kappa ^\steptt)
	   \GammaDist(\beta;   a_\beta  ^\steptt, b_\beta  ^\steptt), \\
	\label{EqTrialDistributionPhi}
	q^\steptt(\bPhi)
	&= \prod_{l=1}^L \NormalDist(\bphi_l; \bmu_{\bphi_l}^\steptt, \bSigma_{\bphi_l}^\steptt).
\end{align}
For (\ref{EqRecurrenceInitiation}) and (\ref{EqRecurrenceRelation}), we update those distributions as follows. First, we compute $q^\steptl(\bmeta)$ using $q^\steptt(\bx, \lambda, \rho, \kappa, \beta, \bPhi)$. Second, we compute $q^\steptl(\bx)$ using $q^\steptl(\bmeta) q^\steptt(\lambda, \rho, \kappa, \beta, \bPhi)$. Finally, we compute $q^\steptl(\lambda, \rho, \kappa, \beta)$ using $q^\steptl(\bx, \bmeta) q^\steptt(\bPhi)$ and $q^\steptl(\bPhi)$ using $q^\steptl(\bx, \bmeta) q^\steptt(\lambda, \rho, \kappa, \beta)$. Here, we simply compute only the parameters of those distributions because we can compute the expectations in (\ref{EqRecurrenceRelation}) analytically by using Taylor approximations in (\ref{EqTaylorApproximation1}), (\ref{EqTaylorApproximation2}), and (\ref{EqTaylorApproximation3}). Specific update equations are described in the Appendix.

For the initial parameters of the trial distributions of $\bmeta$ and $\bx$, we use non-informative values,
\begin{align}
	\bmu_\bmeta^\stepOO \equiv \bm{0},~ \bmu_\bx^\stepOO \equiv \bm{0},~ \bSigma_\bx^\stepOO \equiv \bm{0}.
\end{align}
For the initial parameters for $\lambda$, $\rho$, $\kappa$, $\beta$ and $\bPhi$, we use the same values as their prior's values.

We obtain the well-approximated PM of $\bx$ as $\bmu_\bx^\stepinf$. Realistically, instead of $\bmu_\bx^\stepinf$, we use $\bmu_\bx^\steptl$ when the following convergence conditions hold for $\bmu_\bx^\steptl$ and each $\mu_{\phi_{l,k}}^\steptl$,
\begin{align}
	\frac{1}{N_\bx} \| \bmu_\bx^\steptl - \bmu_\bx^\steptt \|_2^2 &< 10^{-4},\nonumber\\
	\frac{1}{L} \sum_{l=1}^L \frac{(\mu_{\phi_{l,k}}^\steptl-\mu_{\phi_{l,k}}^\steptt )^2}{[\bsigma^2_{\bphi}]_{k}} &< 10^{-4} \quad (k=1,2,3,4),
\end{align}
where we defined $\bsigma^2_{\bphi} \equiv [10^{-3}, 10^{0}, 10^{0}, 10^{-3}]$ as the scaling constant.

\begin{figure*}[tb]
	\centering
	\subfigure[Lena]{\includegraphics[width=30mm]{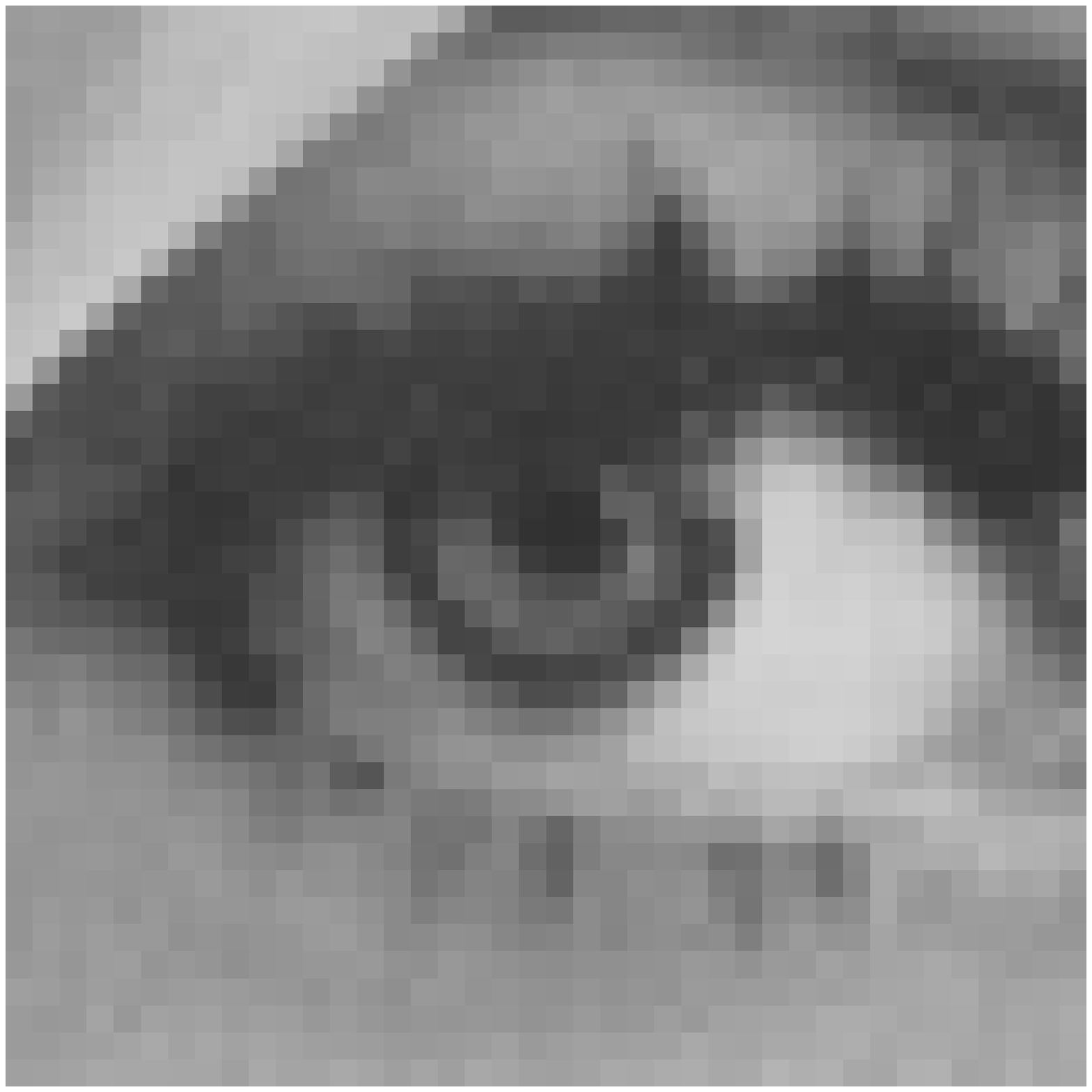}
	\label{OriginalLena}}
	\subfigure[Cameraman]{\includegraphics[width=30mm]{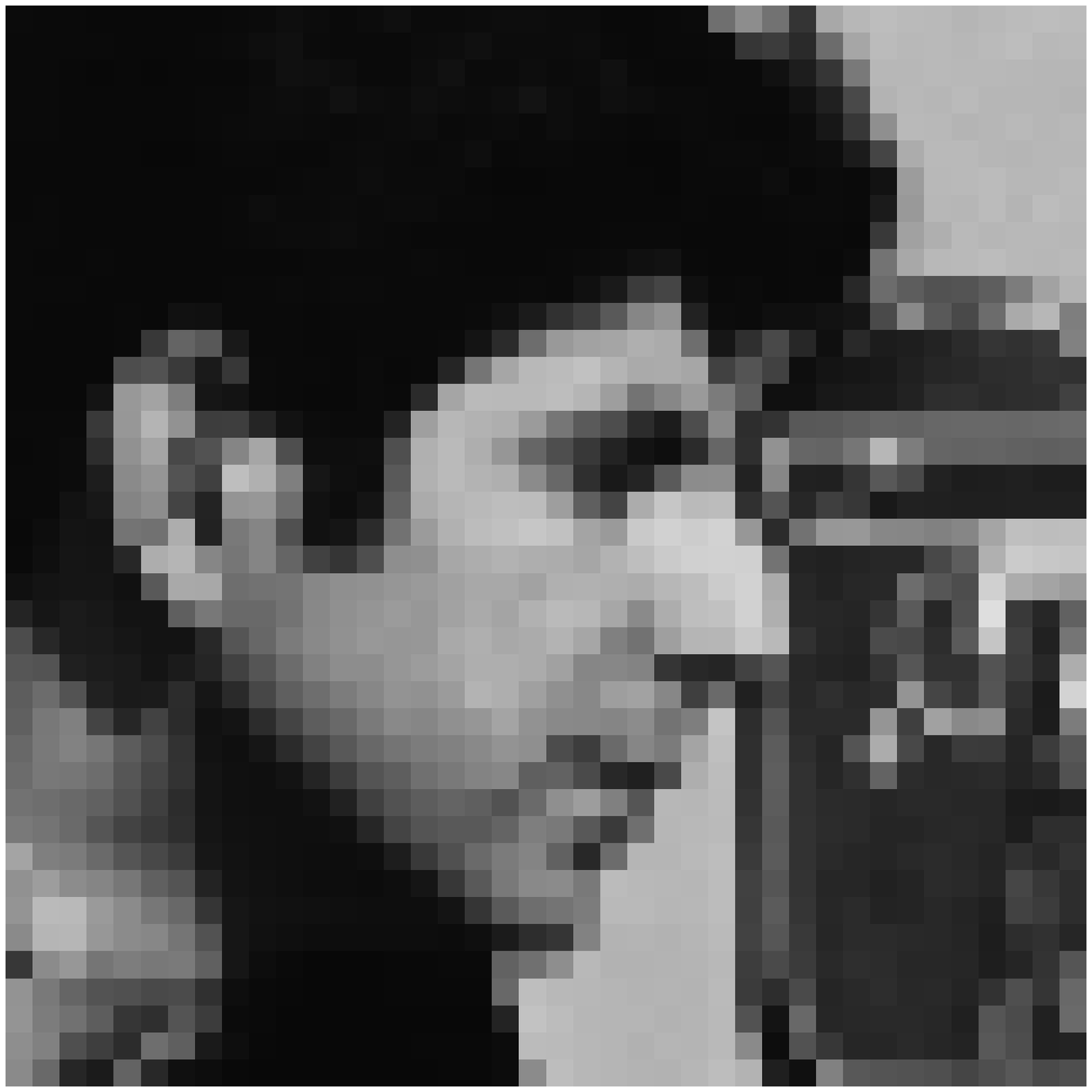}
	\label{OriginalCameraman}}
	\subfigure[Pepper]{\includegraphics[width=30mm]{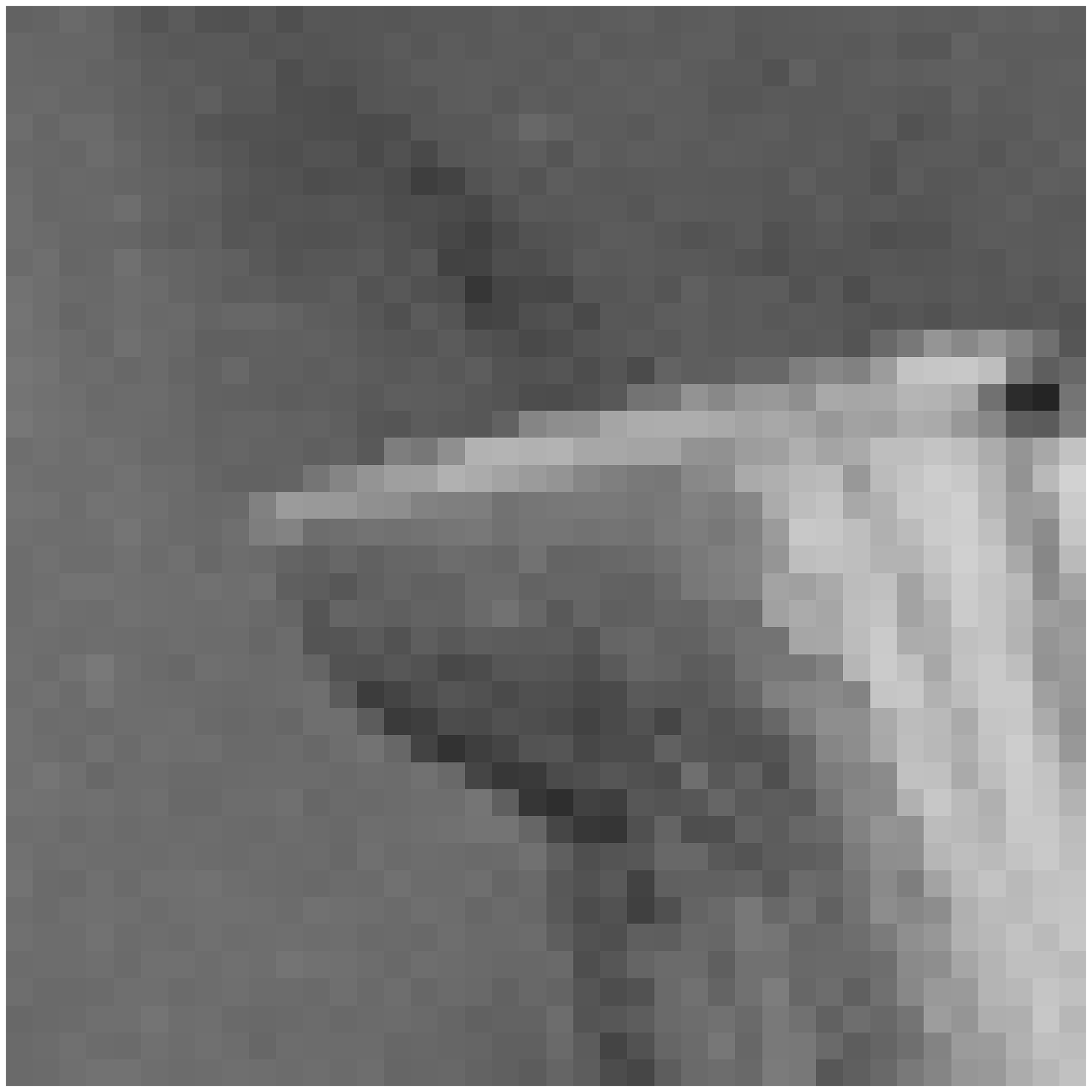}
	\label{OriginalPepper}}
	\subfigure[Clock]{\includegraphics[width=30mm]{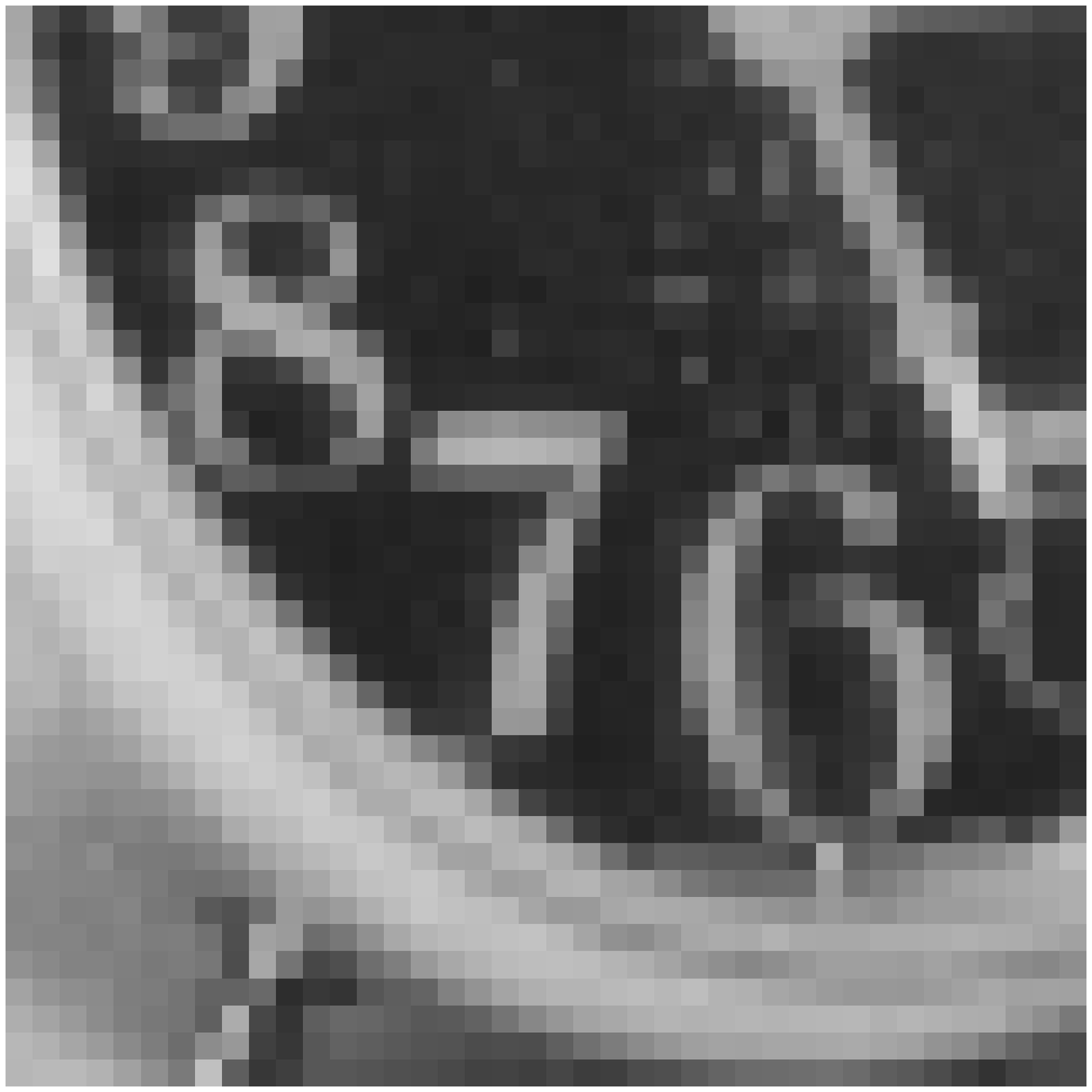}
	\label{OriginalClock}}
	\subfigure[Text]{\includegraphics[width=30mm]{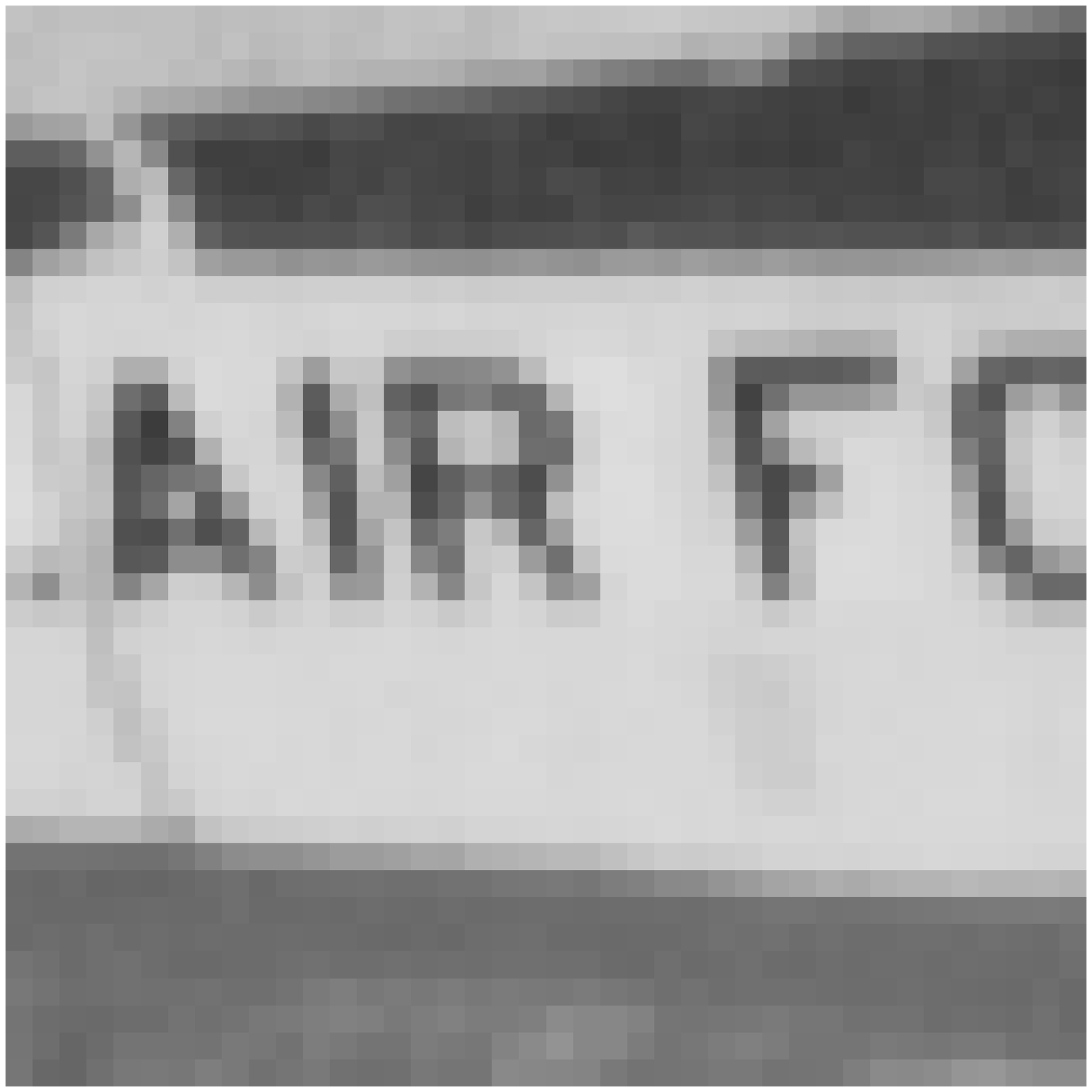}
	\label{OriginalText}}
	\caption{Five original images used in the experiments}
	\label{FigOriginal}
\end{figure*}
\begin{figure*}[tb]
	\centering
	\subfigure[Lena]{\includegraphics[width=30mm]{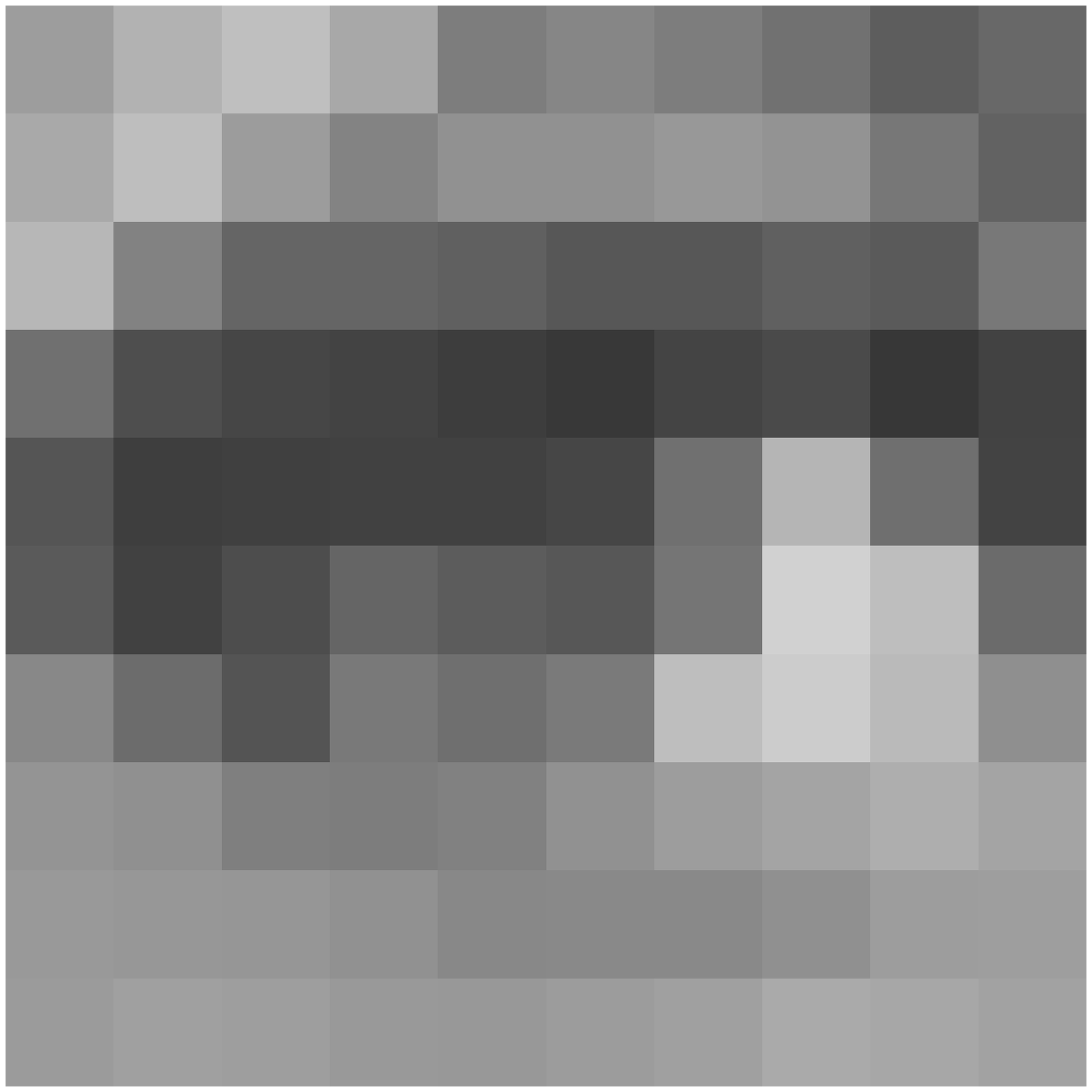}
	\label{ObservedLena}}
	\subfigure[Cameraman]{\includegraphics[width=30mm]{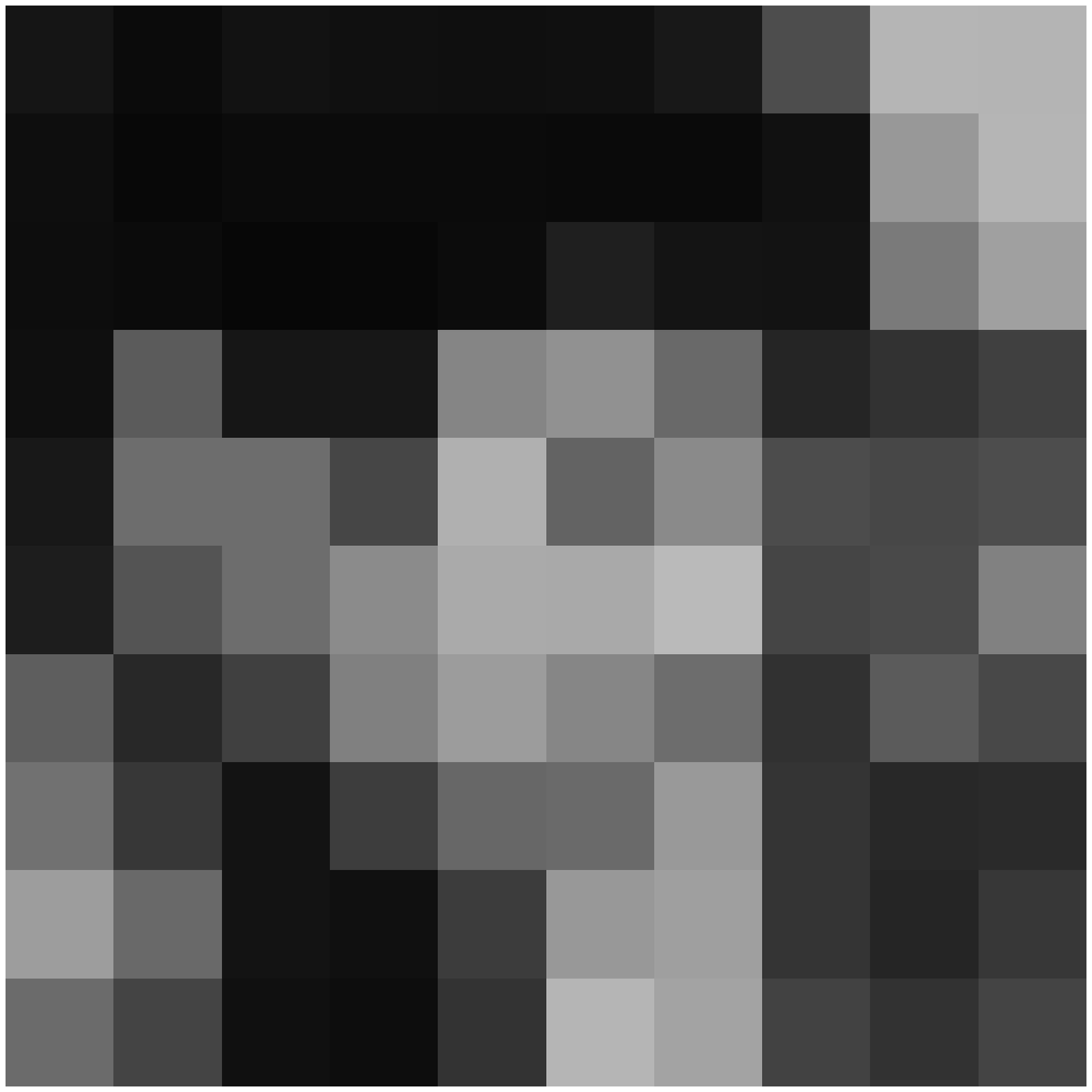}
	\label{ObservedCameraman}}
	\subfigure[Pepper]{\includegraphics[width=30mm]{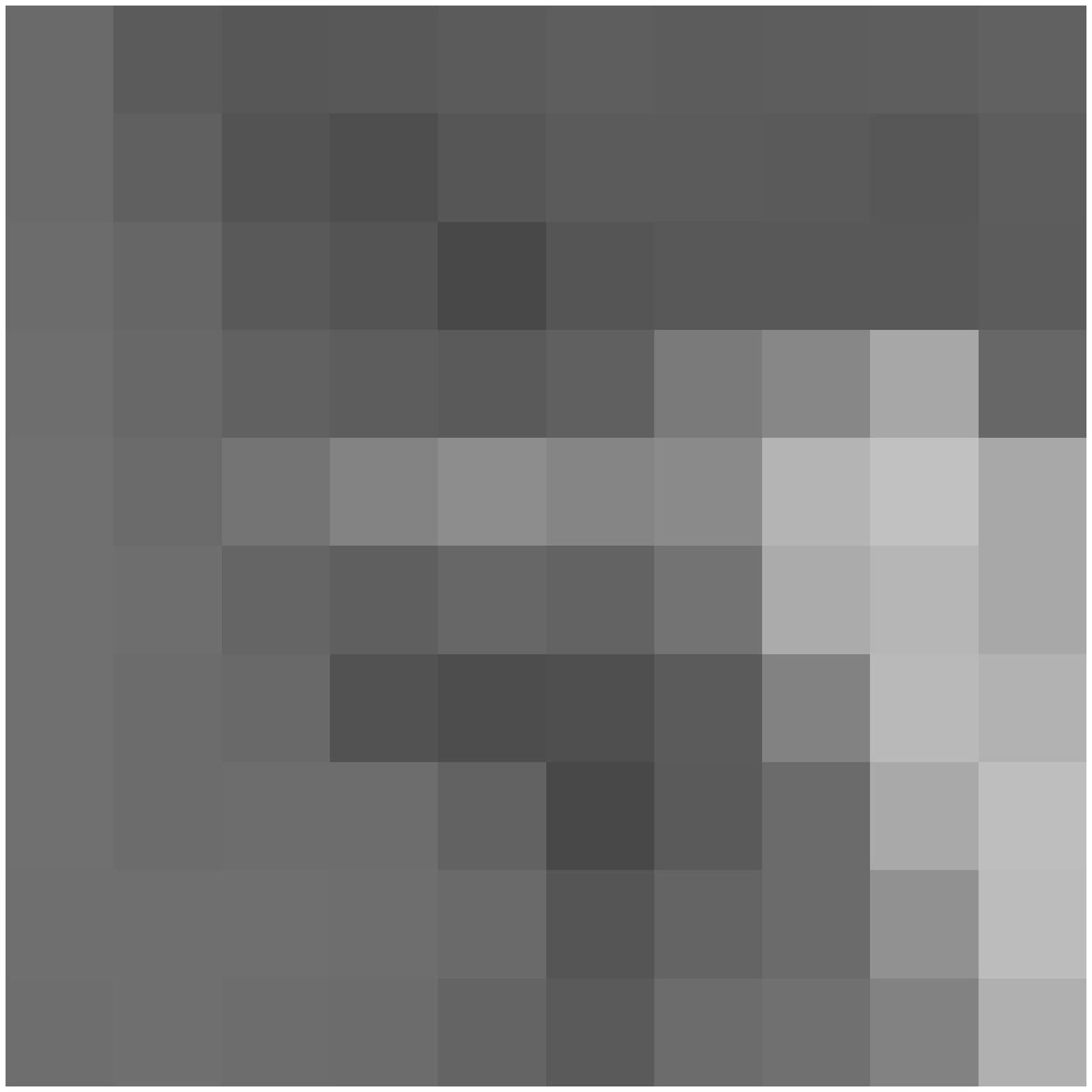}
	\label{ObservedPepper}}
	\subfigure[Clock]{\includegraphics[width=30mm]{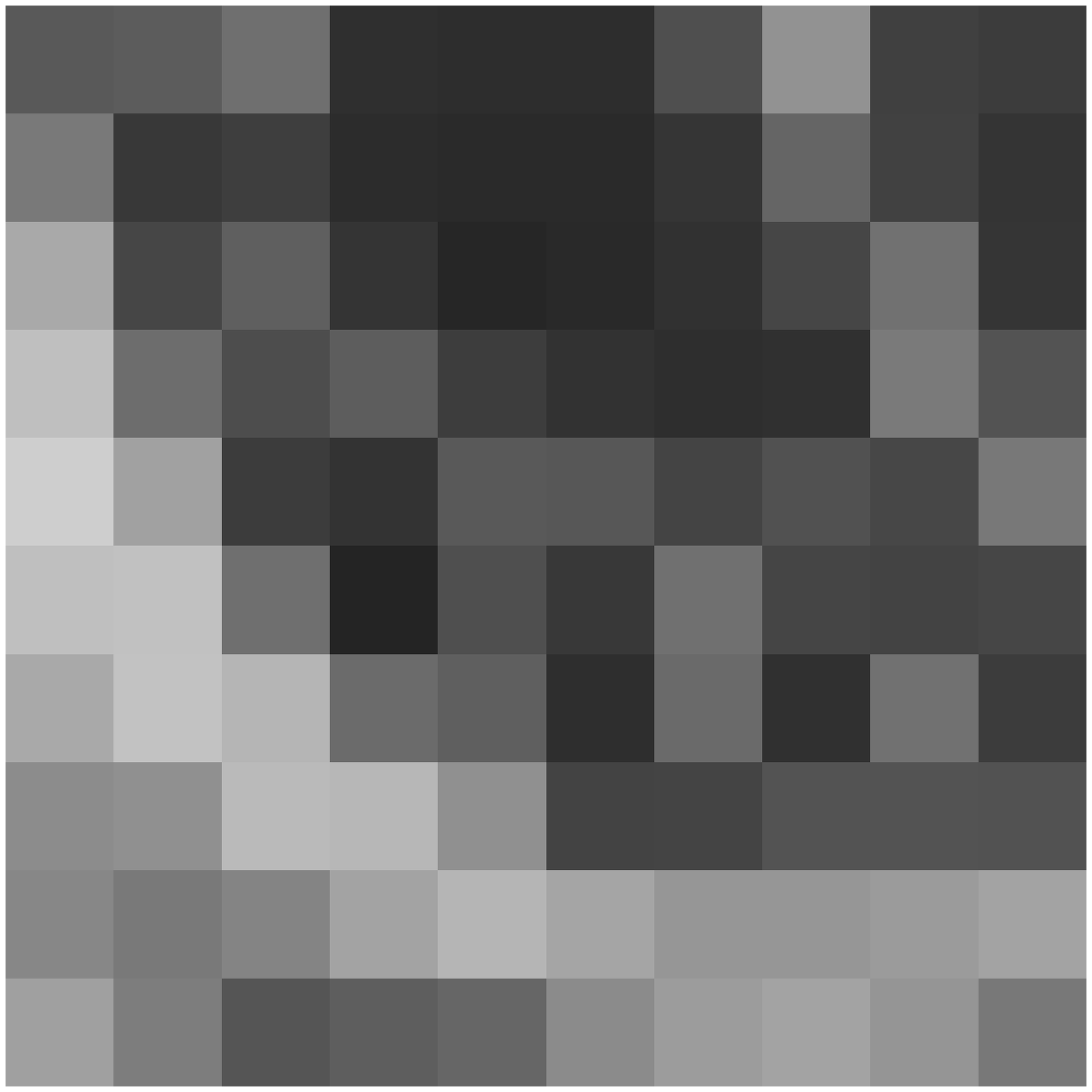}
	\label{ObservedClock}}
	\subfigure[Text]{\includegraphics[width=30mm]{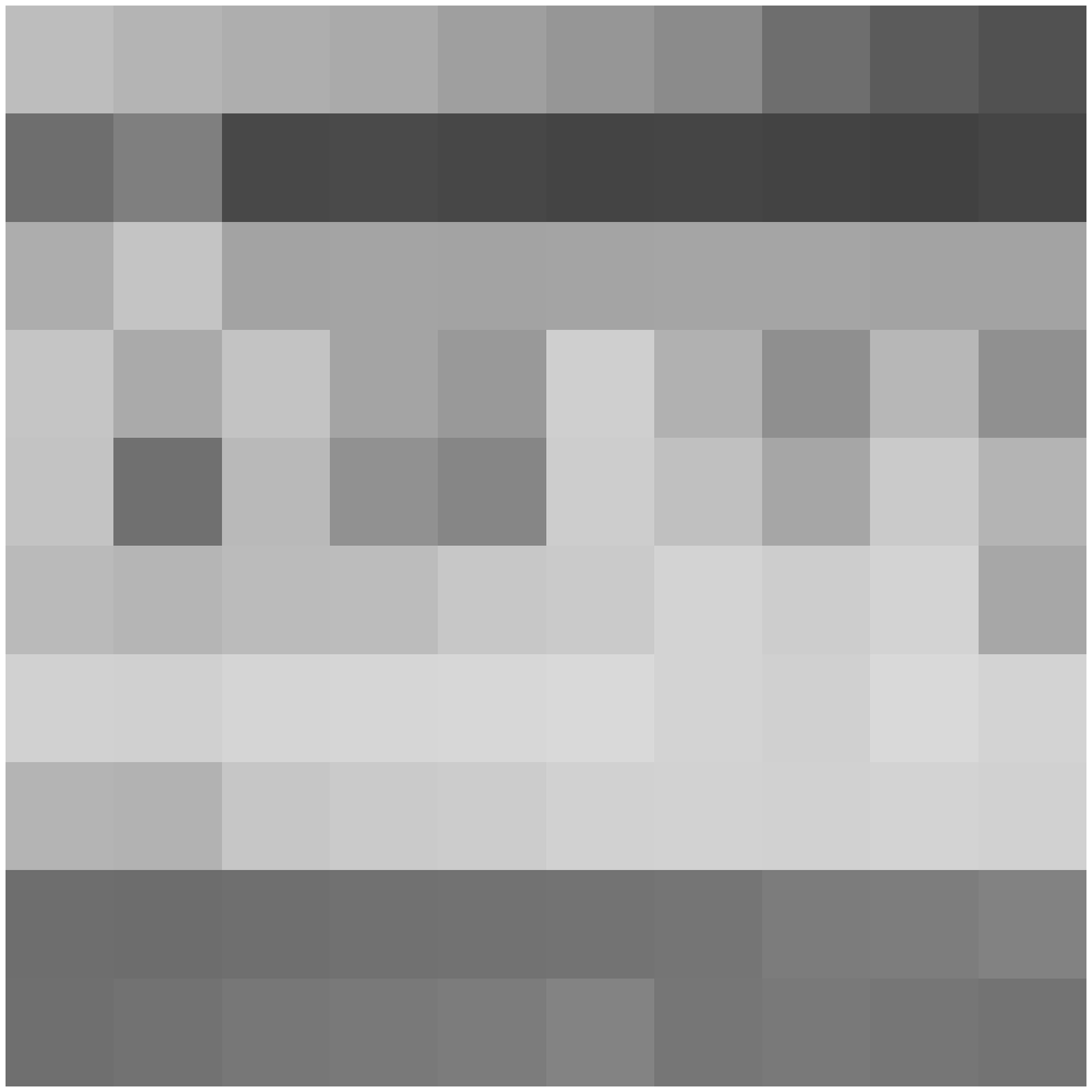}
	\label{ObservedText}}
	\caption{Observed images when warped, blurred, downsampled by an enhancement factor of 4, and noised with SNR$=30$dB AWGN}
	\label{FigObserved}
\end{figure*}
\begin{figure*}[tb]
	\centering
	\subfigure[Lena]{\includegraphics[width=30mm]{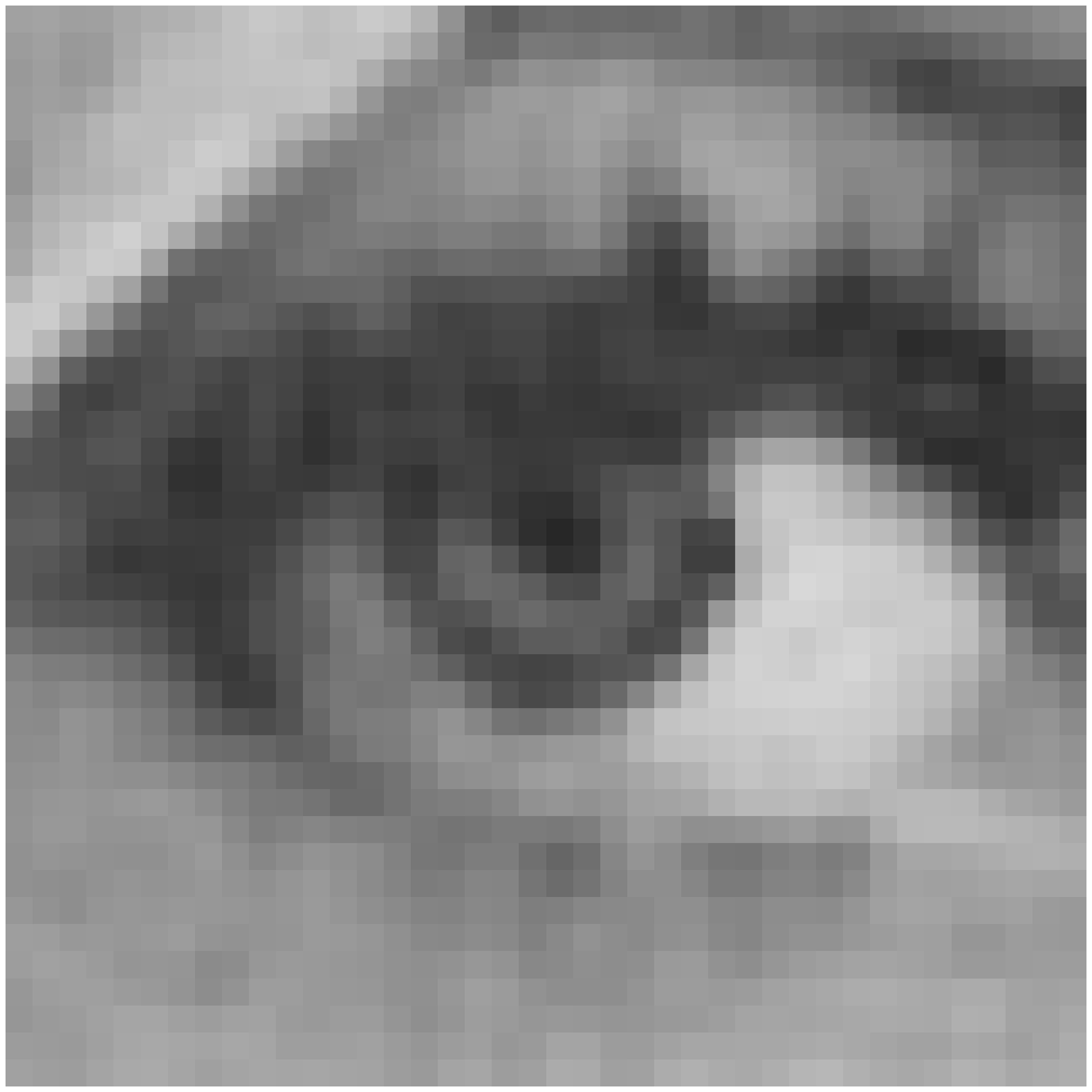}
	\label{InferredLena}}
	\subfigure[Cameraman]{\includegraphics[width=30mm]{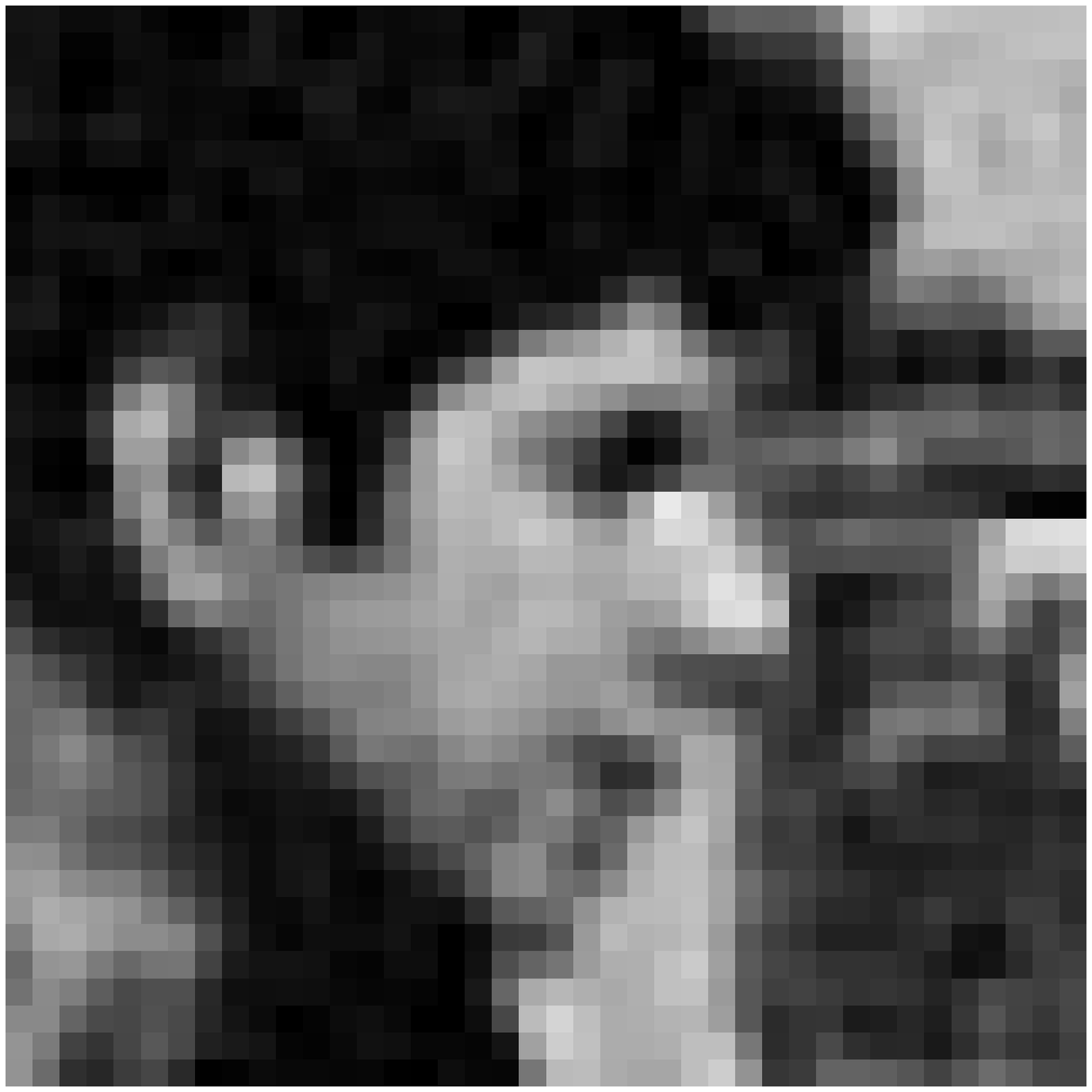}
	\label{InferredCameraman}}
	\subfigure[Pepper]{\includegraphics[width=30mm]{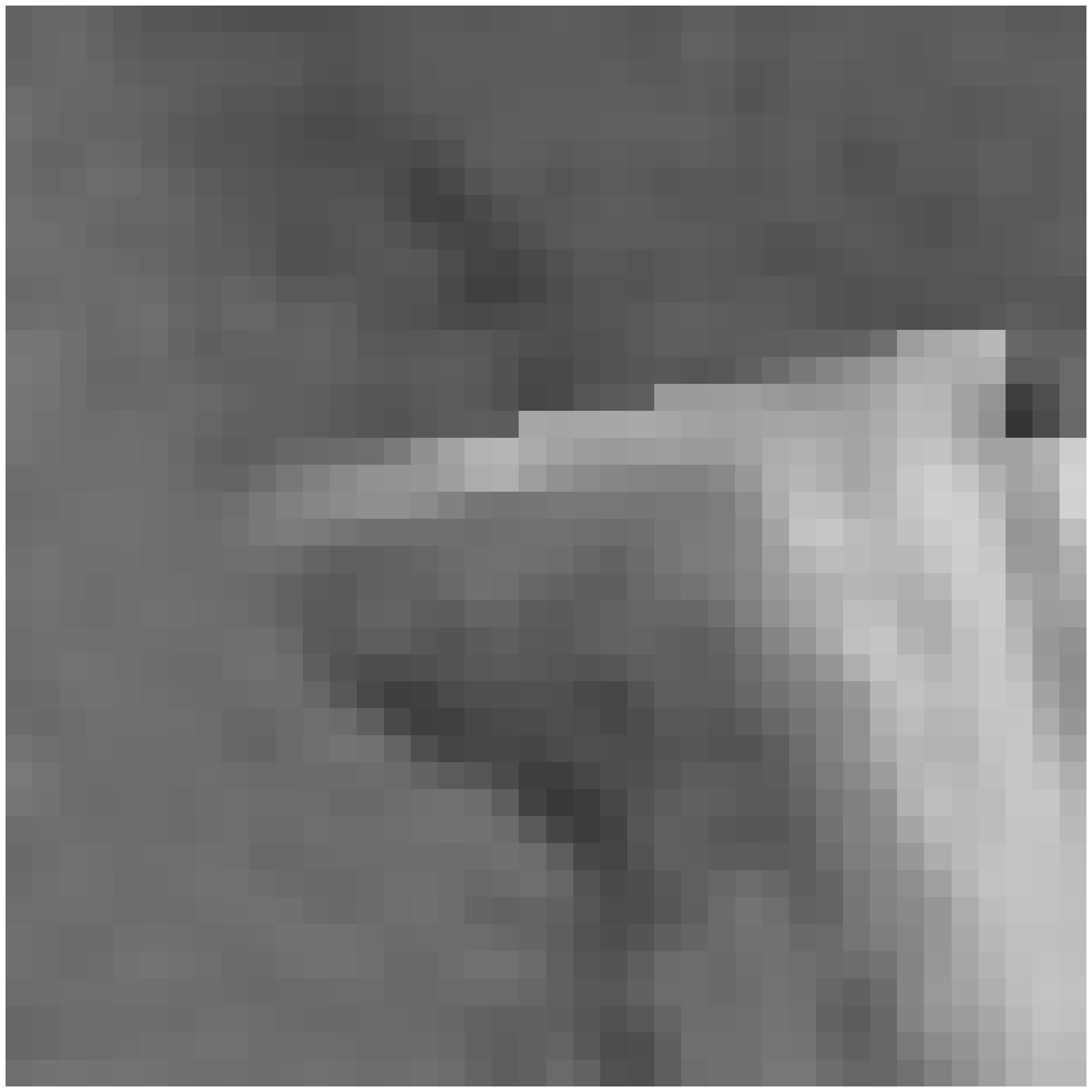}
	\label{InferredPepper}}
	\subfigure[Clock]{\includegraphics[width=30mm]{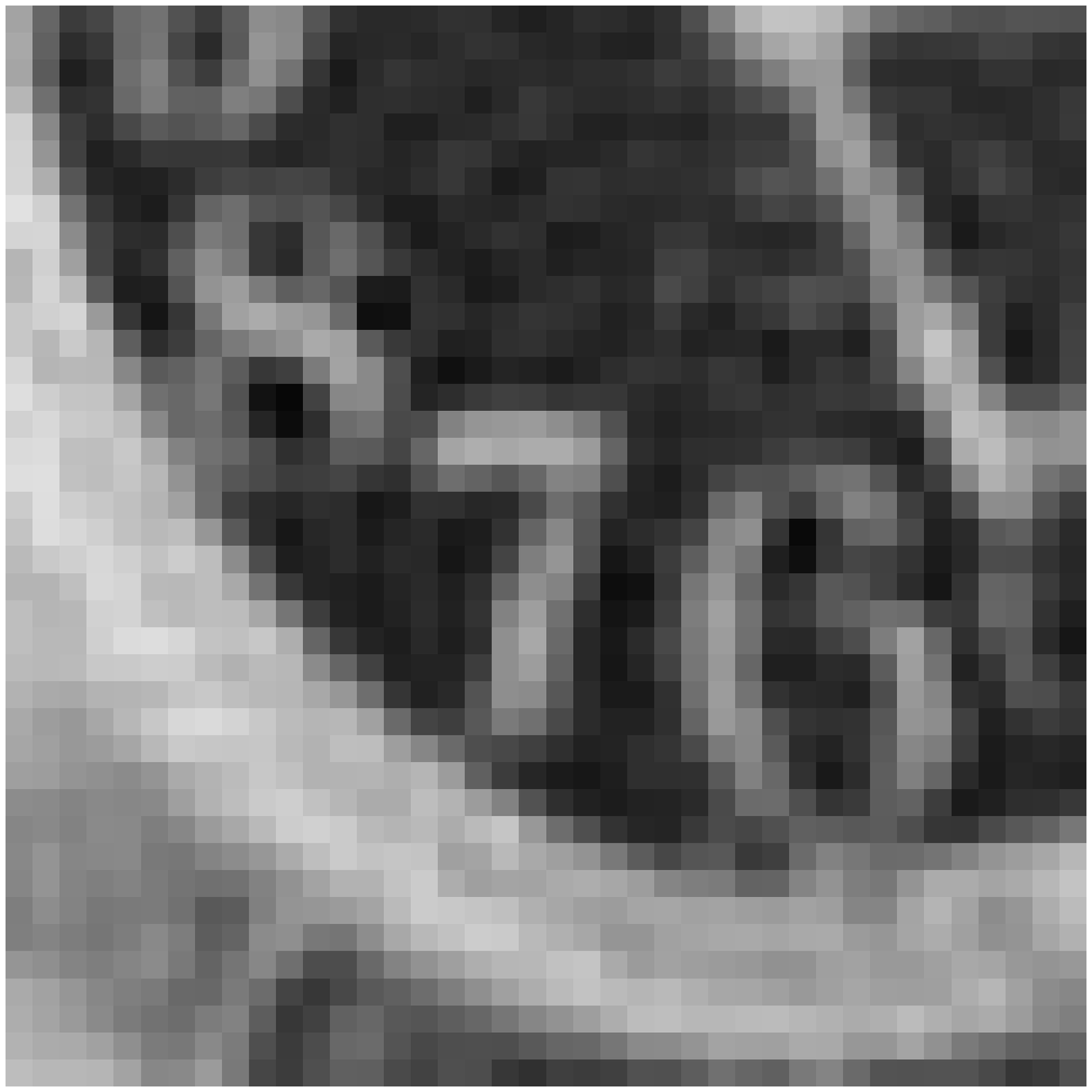}
	\label{InferredClock}}
	\subfigure[Text]{\includegraphics[width=30mm]{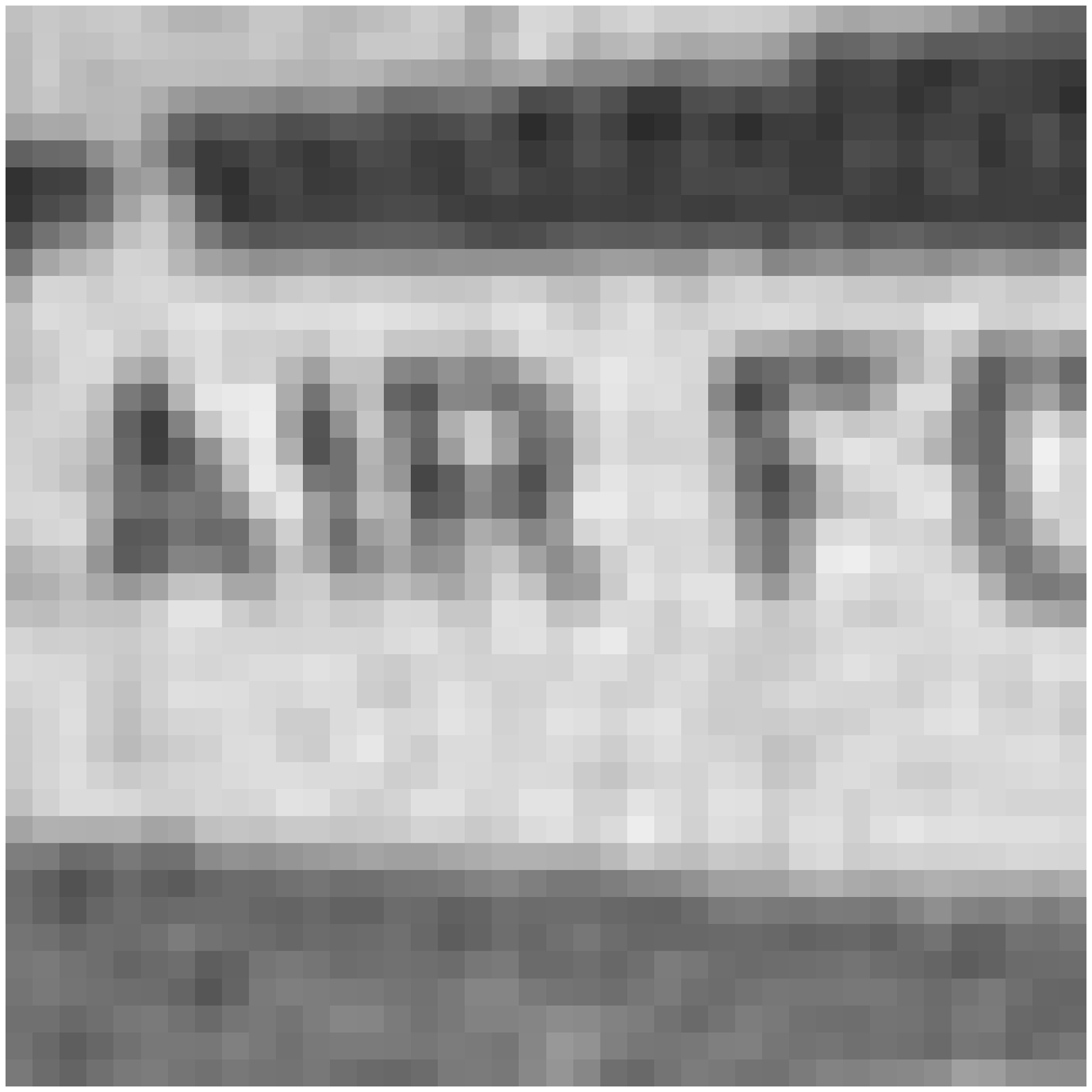}
	\label{InferredText}}
	\caption{Images estimated from Fig. \ref{FigObserved} observed images}
	\label{FigInferred}
\end{figure*}

\section{Experimental Results}
\begin{table*}[tb]
	\centering
	\caption{PSNR of the proposed method (a higher value is better) and ISNRs against three previous methods (a higher value is better) for different images and SNR levels}
	\label{TableResultsPSNR}
	\begin{tabular}{cccccc} 
	\hline\hline
	Image & SNR  & PSNR       & \multicolumn{3}{c}{ISNR} \\\cline{4-6}
	      & [dB] & (proposed) & (vs bilinear) & (vs Kanemura) & (vs Babacan)\\
	\hline
	Lena          & $20$ & $29.22 \pm 0.33$ & $+5.35 \pm 0.33$ & $+0.73 \pm 0.36$ & $-0.11 \pm 0.08$ \\
	              & $25$ & $30.69 \pm 0.25$ & $+6.78 \pm 0.30$ & $+1.04 \pm 0.32$ & $+0.11 \pm 0.11$ \\ 
	              & $30$ & $32.13 \pm 0.36$ & $+8.18 \pm 0.39$ & $+1.60 \pm 0.37$ & $+0.48 \pm 0.24$ \\ 
	\hline
	Cameraman     & $20$ & $21.74 \pm 0.19$ & $+4.11 \pm 0.20$ & $+1.06 \pm 0.36$ & $+0.05 \pm 0.07$ \\
	              & $25$ & $22.68 \pm 0.21$ & $+5.00 \pm 0.23$ & $+1.16 \pm 0.34$ & $-0.06 \pm 0.08$ \\ 
	              & $30$ & $23.72 \pm 0.38$ & $+6.04 \pm 0.39$ & $+1.81 \pm 0.17$ & $+0.03 \pm 0.06$ \\ 
	\hline
	Pepper        & $20$ & $29.71 \pm 0.37$ & $+3.69 \pm 0.35$ & $+0.11 \pm 0.17$ & $+0.28 \pm 0.12$ \\
	              & $25$ & $30.69 \pm 0.28$ & $+4.57 \pm 0.27$ & $+0.28 \pm 0.32$ & $+0.06 \pm 0.16$ \\ 
	              & $30$ & $31.23 \pm 0.42$ & $+5.10 \pm 0.43$ & $+0.81 \pm 0.75$ & $-0.34 \pm 0.48$ \\ 
	\hline
	Clock         & $20$ & $23.27 \pm 0.18$ & $+5.36 \pm 0.19$ & $+1.49 \pm 0.22$ & $+0.11 \pm 0.10$ \\
	              & $25$ & $24.29 \pm 0.22$ & $+6.35 \pm 0.22$ & $+1.77 \pm 0.26$ & $+0.09 \pm 0.08$ \\ 
	              & $30$ & $25.49 \pm 0.50$ & $+7.53 \pm 0.51$ & $+2.49 \pm 0.19$ & $+0.32 \pm 0.13$ \\ 
	\hline
	Text          & $20$ & $24.67 \pm 0.26$ & $+5.83 \pm 0.28$ & $+1.57 \pm 0.20$ & $-0.10 \pm 0.04$ \\
	              & $25$ & $25.87 \pm 0.30$ & $+7.00 \pm 0.33$ & $+1.98 \pm 0.33$ & $-0.03 \pm 0.14$ \\ 
	              & $30$ & $27.26 \pm 0.65$ & $+8.37 \pm 0.67$ & $+3.03 \pm 0.42$ & $+0.18 \pm 0.06$ \\ 
	\hline\hline
	\end{tabular}
\end{table*}
The proposed method was evaluated using five gray-scale images with a size of $40 \times 40$ pixels, as shown in Fig. \ref{FigOriginal}. From each image, $L=10$ images with a size of $10 \times 10$ pixels were created by using (\ref{EqImageObservationProcess}), (\ref{EqObservationModel}) with the settings of the parameters $\alpha$, $\bPhi$, and $\beta$ as the following. The resolution enhancement factor $\alpha$ was 4. The transformation parameter $\bPhi$ was randomly created according to the prior distribution in (\ref{EqPriorDistributionPhi}). The noise level parameter $\beta$ was set for signal-to-noise ratios (SNR) of $20$, $25$, and $30$ dB for each image. Samples of the created images are shown in Fig. \ref{FigObserved}.

Figure \ref{FigInferred} shows the images estimated under SNR$=30$dB. The resolution of each image appeared to be better than the corresponding observed image in Fig. \ref{FigObserved}.

Table \ref{TableResultsPSNR} lists the quantitative results compared to those from the methods of bilinear interpolation, Kanemura \textit{et al.} \cite{Kanemura2007}, and Babacan \textit{et al.} \cite{Babacan2011}. Note that we added a slight modification to these methods because they employ slightly different models. For example, the original method \cite{Babacan2011} assumes the blurring parameter $\gamma$ is known, so we set $\gamma$ as the mean value of the true distribution for this method. Also, we introduced a strong prior for $\lambda$ in the Kanemura method \cite{Kanemura2007} in contrast to the original method, because this parameter sometimes becomes negative. We evaluated the results with regard to the expectation and the standard deviation of the improvement in signal-to-noise ratio (ISNR) over $10$ experiments on each image and for each SNR. ISNR is the relative PSNR defined as
\begin{align}
	\mathrm{ISNR} \equiv \mathrm{PSNR}(\hbx; \bx) - \mathrm{PSNR}(\tbx; \bx),
\end{align}
where $\bx$ is the true HR image, $\hbx$ is the image estimated by the proposed method, and $\tbx$ is the image estimated by the compared method. A higher ISNR value means better improvement of the estimate against the estimate of the compared method.
We see that the ISNRs of the proposed method were mostly higher than those of the other methods, except for the comparison with the Babacan's method in Pepper image.

Table \ref{TableResultsRegistration} lists the root mean square errors (RMSE) of our method and the other methods. To evaluate the estimated registration parameters, we took the RMSEs over $50$ experiments ($10$ experiments $\times$ 5 images) for each noise level. Of course, a lower RMSE value means a better estimate. We see that the RMSEs of the proposed method were mostly higher than those of the other methods.

The calculation times of the proposed method was about $10$ minutes on an Intel Core i7 2600 processor. The proposed method was a little slower than the method of Babacan \textit{et al.} \cite{Babacan2011} and a little faster than the method of Kanemura \textit{et al.} \cite{Kanemura2007}.

\section{Discussion}
With regard to the observation model, we used a linear transformation and AWGN. The use of the linear transformation model is advantageous since an arbitrary transformation matrix $\bW(\bphi_l)$ can be employed because of the Taylor approximation. The transformation matrix can be constructed by multiplying three matrices: the warping, blurring, and downsampling matrices \cite{Babacan2011}. A disadvantage of this is that sub-pixel errors might accumulate. We prefer matrix construction via a continuous function \cite{Tipping2003}. We improved the construction by introducing an elliptic theta function for the normalizing constant in (\ref{EqTransformationMatrixEllipticTheta}). This normalizing constant provides fair pixel weights for both marginal and central areas of the HR image and faithfully represents the Gaussian PSF.

With regard to the HR image prior, we used a causal type of prior, which was first introduced by Kanemura \textit{et al.} \cite{Kanemura2007,Kanemura2009}. The microstate energy function, or equivalently, the Hamiltonian, -based compound MRF prior of (\ref{EqPriorDistributionAnotherXEta}), offers the advantage of easy construction, but it usually has an exponential calculation cost, $\calO(2^{N_\bmeta})$, for the normalizing constant or, equivalently, the partition function, and this is an obstacle to direct calculation of the PM solution. The MAP solution has been used in work elsewhere because it is not affected by the normalizing constant. In contrast, the introduced causal type of prior of (\ref{EqPriorDistributionXEta}) has only a polynomial calculation cost $\calO(N_\bx^3)$, which enables us to successfully apply the variational Bayes method to this problem.

\begin{figure*}[tb]
	\centering
	\subfigure[Lambda]{\includegraphics[width=70mm]{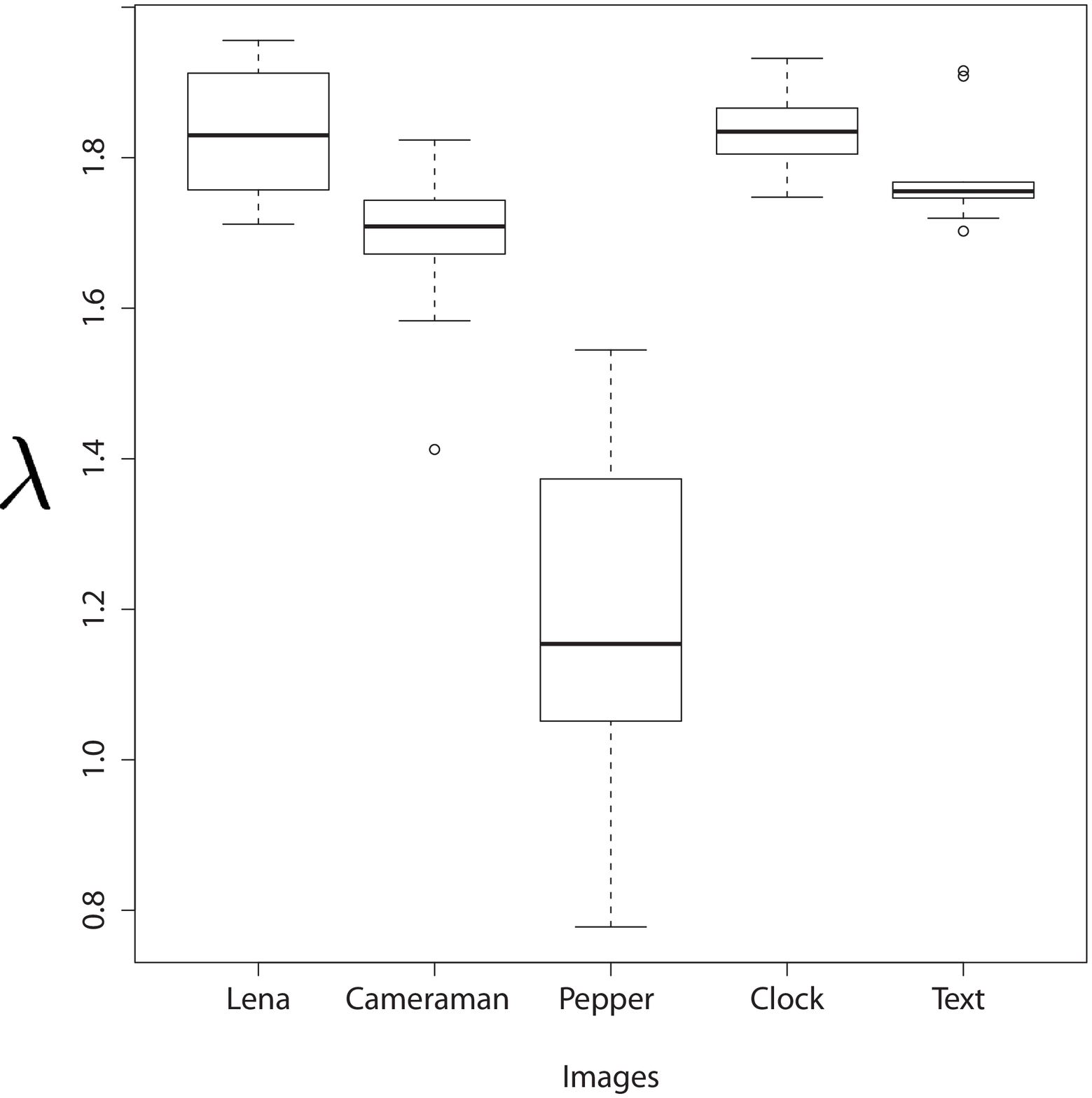}
\label{Lambda}}\hspace{40pt}
	\subfigure[Rho]{\includegraphics[width=70mm]{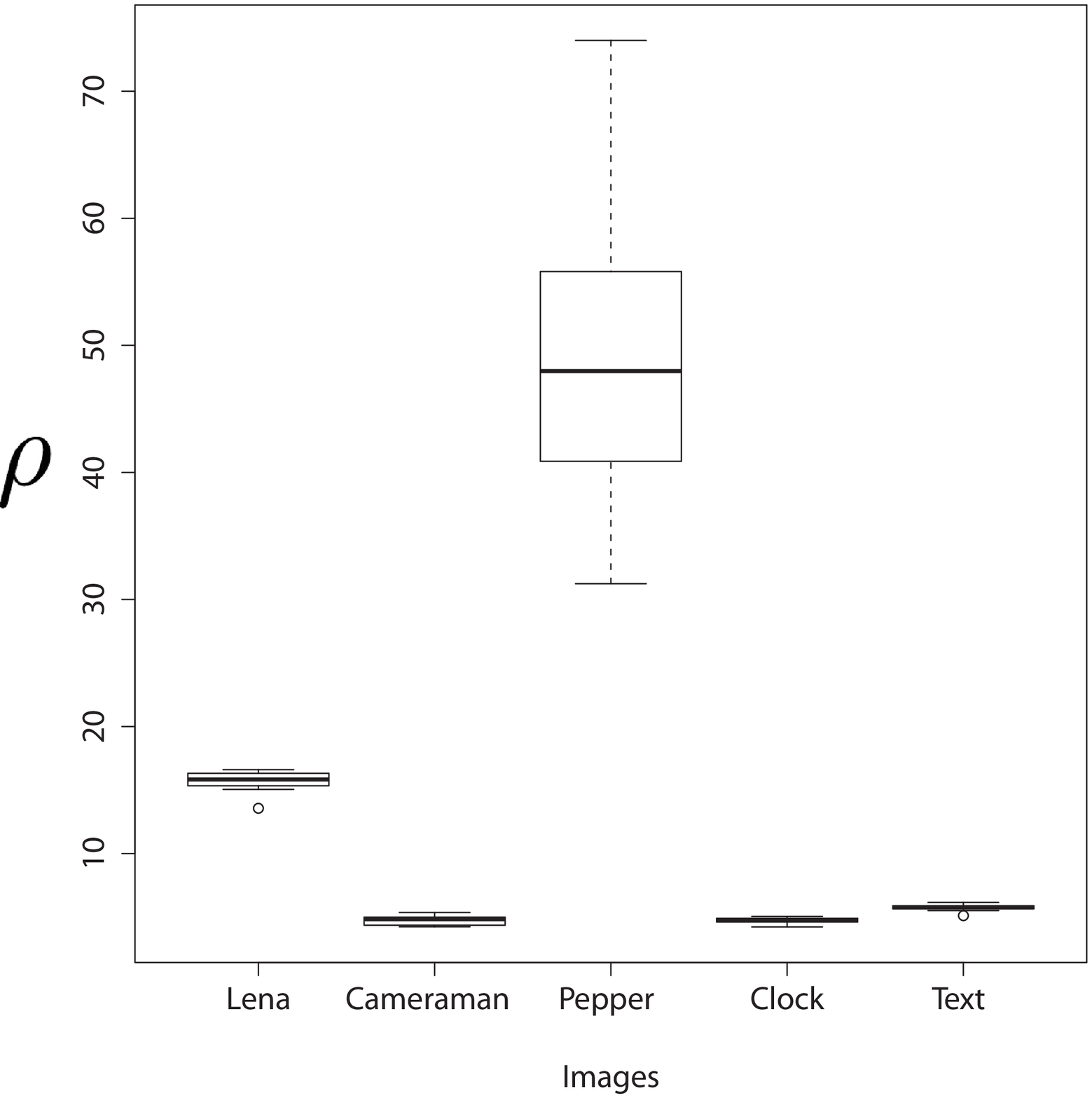}
\label{Rho}}
	\subfigure[Kappa]{\includegraphics[width=70mm]{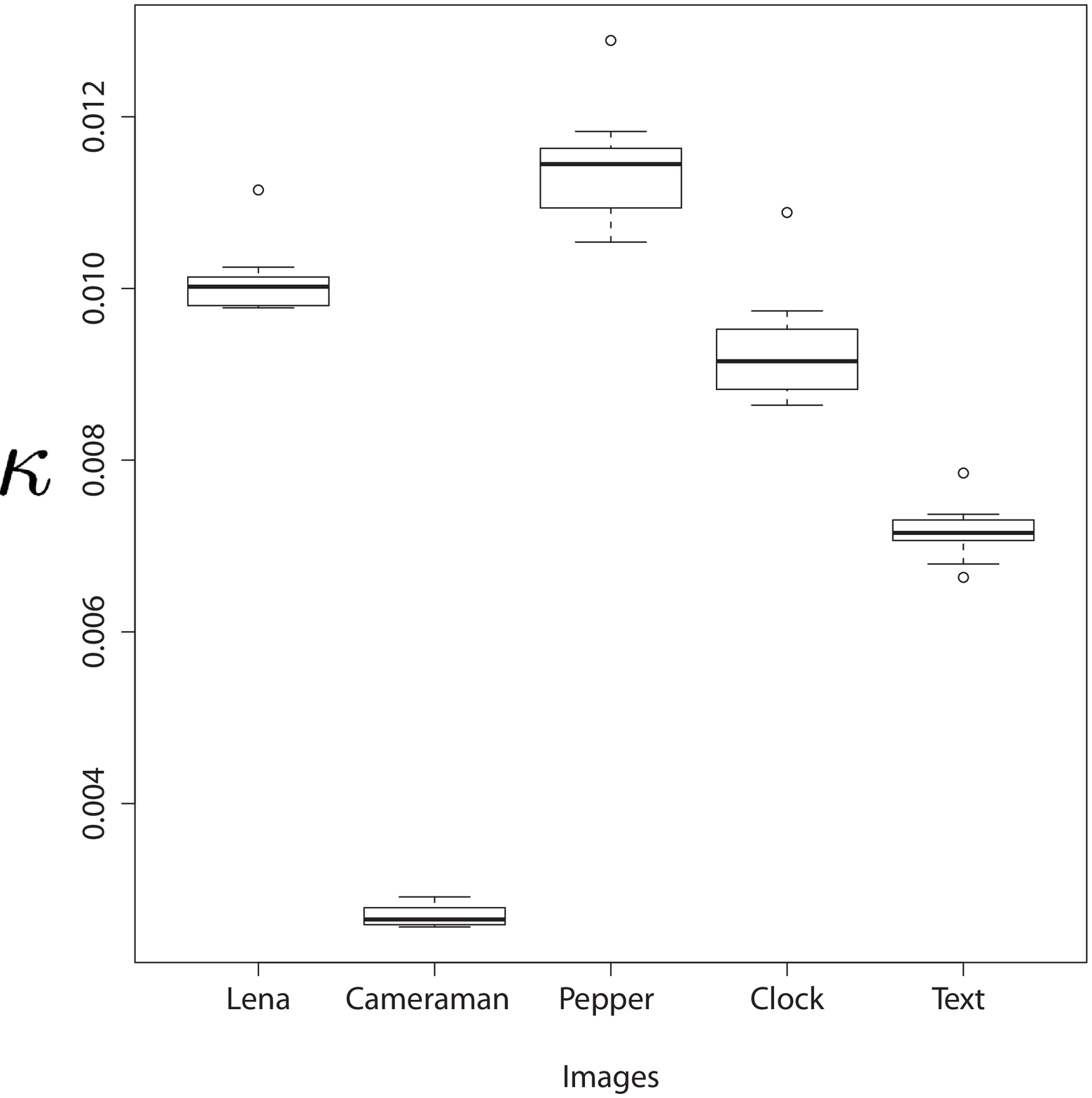}
\label{Kappa}}\hspace{40pt}
	\subfigure[Beta]{\includegraphics[width=70mm]{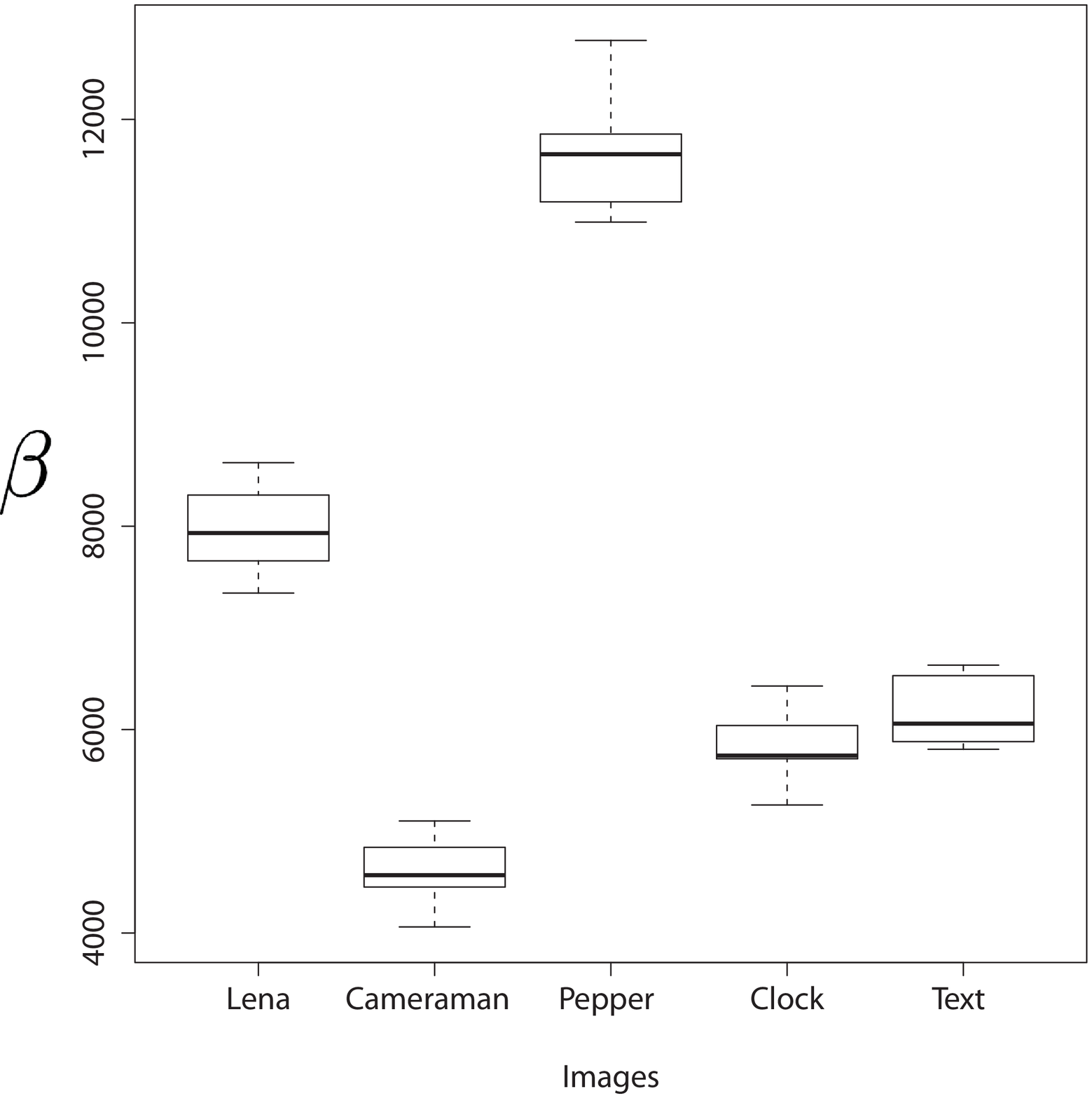}
\label{Beta}}
	\caption{Box and whisker plot of the PM for each hyperparameter, $\lambda$, $\rho$, $\kappa$, and $\beta$, and image under SNR$=30$dB noise}
	\label{FigRho}
\end{figure*}

With regard to the hyperparameter priors, we also improved the existing method. As the edge penalty parameter $\lambda$, Kanemura \textit{et al.} \cite{Kanemura2007} implicitly assumed $\lambda \in \mathbb{R}$, which leads to a negative $\lambda$ and consequently results in an edge-strewn image. We assumed $\lambda > 0$ by setting its prior according to a gamma distribution, resulting in an appropriate inference. As the smoothness parameter $\rho$, they practically fixed the value of $\rho$ with a strongly informative prior. We chose a non-informative prior for $\rho$. We show the box and whisker plot of the PM for each hyperparameter over $10$ experiments on each image under SNR$=30$dB noise in Fig. \ref{FigRho}. As can be seen, the inferred value of the PM of $\rho$ showed wide variation, with an approximately 10-fold maximum-to-minimum ratio, depending on the original image. This result can be interpreted as meaning it is worth inferring $\rho$ in each HR image. Furthermore, $\lambda$ and $\kappa$ respectively showed approximately 2-fold and 4-fold ranges of variation. Regarding the contrast parameter $\kappa$, they assumed $\kappa \equiv 0$, which leads to $|\bA| = 0$, and this results in an improper normalizing constant. While we assume $\kappa > 0$, which leads to a proper normalizing constant, we can consequently take the term of $\ln|\bA|$ into account in the update equations of the variational Bayes. 

With regard to the prior distribution for the blurring parameter $\gamma$, we used a Gaussian distribution even though $\gamma$ is a positive real number. This is because we selected a simpler expression. We tried using the prior of the gamma distribution as $\gamma$, but the improvement was small. One disadvantage of this model is that a non-informative setting for this prior may lead to a nonsense result where the inferred $\gamma$ is negative. Moreover, we employed a somewhat informative prior for $\gamma$. This is because the blurring parameter $\gamma$ and smoothness hyperparameter $\rho$ are somewhat complementary. This means that simultaneous estimation of $\gamma$ and $\rho$ is difficult. Tipping \textit{et al.} \cite{Tipping2003} and Kanemura \textit{et al.} \cite{Kanemura2007} fixed $\rho$, and Babacan \textit{et al.} \cite{Villena2010} fixed $\gamma$.

With regard to the estimator, we logically derived the optimal estimator PM from the objective function of the L2-norm-based PSNR. The widely used joint MAP estimator can be considered the optimal estimator for the all-or-none type objective function,
\begin{align}
	\argmax_\hbz \left\langle \delta(\hbz - \bz) \right\rangle_{p(\bz | \bY)} = \argmax_\bz p(\bz | \bY),
\end{align}
where $\delta$ is the Dirac delta or Kronecker delta function. Generally, this type of objective function is nonsensical for continuous variables because it is measure zero. If all the random variables in the posterior distribution are discrete, or if we can assume some smoothness of the posterior distribution, a joint MAP solution will have meaning. Instead of the L2-norm-based objective function of PSNR, the L1-norm (mean absolute error) -based PSNR is sometimes employed. In such cases, the median of the posterior distribution is generally the optimal estimator. In the case of the marginalized ML, or equivalently, type-II ML or empirical Bayes, for example, the registration parameters and other hyperparameters are firstly inferred as:
\begin{align}
	[ \hlambda, \hrho, \hkappa, \hbeta, \hbPhi ] \equiv \argmax_{\lambda, \rho, \kappa, \beta, \bPhi} p(\bY | \lambda, \rho, \kappa, \beta, \bPhi).
\end{align}
If these parameters have priors, such a method is called marginalized MAP. The HR image and sometimes the edge information are then inferred to as MAP,
\begin{align}
	\hbx \equiv \argmax_\bx \max_{\bmeta} p(\bx, \bmeta | \bY, \hlambda, \hrho, \hkappa, \hbeta, \hbPhi),
\end{align}
or PM. For such a two-step inference, it is difficult to calculate back the objective function.

With regard to the Taylor approximation for the transformation matrix $\bW(\bphi_l)$, we used the first-order approximation in (\ref{EqTaylorApproximation1}) because it is more stable than the second-order approximation. This first-order approximation was proposed by Villena \textit{et al.} \cite{Villena2010}. The second-order approximation was proposed by Pickup \textit{et al.} \cite{Pickup2007}, and they obtained good results. We also tried the second-order approximation, but it sometimes made the algorithm unstable because it sometimes failed to produce a positive definite matrix for the covariance matrix $\bSigma_\bx$.

With regard to the Taylor approximation for $\ln|\bA(\bmeta, \rho, \kappa)|$ and $\ln \logistic(\lambda)$, we introduced the first-order approximation around $[\bmeta, \ln \rho, \ln \kappa] = [\bmu_\bmeta^\steptt, \ln \mu_\rho^\steptt, \ln \mu_\kappa^\steptt]$ and $\ln \lambda = \ln \mu_\lambda^\steptt$, respectively, in (\ref{EqTaylorApproximation2}) and (\ref{EqTaylorApproximation3}). Note that the Taylor expansion not with respect to $\rho, \kappa, \lambda$, but with respect to $\ln \rho, \ln \kappa, \ln \lambda$ is our key idea to solve the conjugate prior problem. Indeed, we could successfully derive the terms originating from $\ln|\bA|$ in update equations ((\ref{EqUpdateMuEtaA}), (\ref{EqUpdateARho}), and (\ref{EqUpdateAKappa}) in Appendix). Kanemura \textit{et al.} \cite{Kanemura2007,Kanemura2009} ignored the term of $\ln|\bA|$ because of the high calculation cost, and this would result in less accurate inference. As for $\bmeta$, we implicitly assumed that $\bmeta$ is not a binary vector but a continuous vector and did the differentiation. This assumption is based on (\ref{EqAMatrix}). If we make another assumption -- i.e., replacement of $\eta_{i,j}$ with $\eta_{i,j}^2$ in (\ref{EqAMatrix}) -- (\ref{EqAMatrix}) has the same meaning, but the result of the Taylor approximation will differ from the current form.

With regard to the experimental results, the proposed method outperforms the other methods in terms of the ISNR for most images and noise levels. Moreover, its estimation of the registration parameters was more accurate than the other methods were for most conditions. Therefore, we conclude the proposed method is on the whole superior to the other methods. Compared with bilinear interpolation and Kanemura's method, the superiority of the proposed method was clear. Compared with the Babacan's method, the superiority of the proposed method was rather slight. Especially, in the case of the Pepper image in $30$ dB noise, the porposed method was worse than the Babacan's method. This inferiority is considered to be caused by unstable estimation of $\gamma$ and $\rho$, where Babacan's method fixed the value of $\gamma$ to the true expected value in our implementation. Intuitively, the Pepper image is smoother than the other images and has fewer edges. Therefore, this feature is considered to be less preferable for complementary parameters of $\gamma$ and $\rho$.

With regard to the calculation cost, the proposed algorithm requires $\calO(N_\bx^3)$. This calculation cost order is given by two matrix inversions: $\bSigma_{\bx}^\steptl$ in (\ref{EqUpdateSigmaX}) and $\bA$ in (\ref{EqUpdateMuEtaA}) and (\ref{EqUpdateAKappa}) (see Appendix). We found that a simple approximation such as considering all the off-diagonal elements to be zero reduces the calculation time but obviously degrades accuracy. We hope to solve this problem in our future work.

\begin{table}[tb]
	\centering
	\caption{RMSEs of registration parameters (a lower value is better) for different SNR levels}
	\label{TableResultsRegistration}
	\begin{tabular}{ccccc} 
	\hline\hline
	parameter & SNR  & \multicolumn{3}{c}{RMSE} \\\cline{3-5}
	          & [dB] & (proposed) & (Kanemura) & (Babacan) \\
	\hline
	$\theta$      & $20$ & $0.006$ & $0.006$ & $0.006$\\
	              & $25$ & $0.004$ & $0.004$ & $0.004$\\
	              & $30$ & $0.002$ & $0.003$ & $0.003$\\
	\hline
	$[\vec{o}]_h$ & $20$ & $0.094$ & $0.095$ & $0.094$\\
	              & $25$ & $0.054$ & $0.059$ & $0.056$\\
	              & $30$ & $0.041$ & $0.060$ & $0.046$\\
	\hline
	$[\vec{o}]_v$ & $20$ & $0.074$ & $0.073$ & $0.076$\\
	              & $25$ & $0.044$ & $0.052$ & $0.047$\\
	              & $30$ & $0.037$ & $0.044$ & $0.036$\\
	\hline
	$\gamma$      & $20$ & $0.031$ & $0.033$ & ---\\
	              & $25$ & $0.025$ & $0.030$ & ---\\
	              & $30$ & $0.028$ & $0.028$ & ---\\
	\hline\hline
	\end{tabular}
\end{table}
\section{Conclusion}
In this paper, we proposed a Bayesian image super-resolution (SR) method with a causal Gaussian Markov random field (MRF) prior. We improved existing models with respect to three points: 1) the combined transformation model through a preferable normalization term using the elliptic theta function, 2) the causal Gaussian MRF model through introduction of a contrast parameter $\kappa$, which provides an effective normalizing constant including $\ln|\bA|$, and 3) the hyperparameter prior model through application of a gamma distribution for the edge penalty parameter $\lambda$, which prevents an unfavorable edge-strewn image. We then logically derived the optimal estimator, that is, not the joint maximum a posteriori (MAP) or marginalized maximum likelihood (ML) but the posterior mean (PM), from the objective function of the L2-norm (mean square error) -based peak signal-to-noise ratio (PSNR). The estimator is numerically determined by using variational Bayes. We solved the conjugate prior problem in variational Bayes by introducing three Taylor approximations. Other than these approximations, we did not use any approximations such as ignoring the term $\ln|\bA|$. Experimental results showed that the proposed method is mostly superior to existing methods in accuracy.

\appendix
Here, we show the details of the variational Bayes' update equations in Section IV-C.

The mean values of the hyperparameters $\lambda, \rho, \kappa,\beta$ over the trial distributions $q^\steptt(\lambda, \rho, \kappa, \beta)$ are given by
\begin{align}
	\mu_\lambda^\steptt = \frac{a_\lambda^\steptt}{b_\lambda^\steptt}, ~
	\mu_\rho   ^\steptt = \frac{a_\rho   ^\steptt}{b_\rho   ^\steptt}, ~
	\mu_\kappa ^\steptt = \frac{a_\kappa ^\steptt}{b_\kappa ^\steptt}, ~
	\mu_\beta  ^\steptt = \frac{a_\beta  ^\steptt}{b_\beta  ^\steptt}.
\end{align}

The update equation of $\bmeta$ is given as
\begin{align}
	q^\steptl(\bmeta)
	&\propto \exp \left\langle \ln p(\bz | \bY) \right\rangle
	_{q^\steptt(\bx, \lambda, \rho, \kappa, \beta, \bPhi)} \nonumber\\
	&\propto \exp \Bigg( \sum_{i \sim j} \left\{ c^\steptt_\lambda - \frac{\mu_\rho^\steptt}{2} \tr \bC_\bx^\steptt \bM_{i,j} \right\} \eta_{i,j} \nonumber\\
	&~~ + \frac{1}{2} \left\langle \ln| \bA(\bmeta, \rho, \kappa) | \right\rangle_{q^\steptt(\rho, \kappa)} \Bigg),
\end{align}
where
\begin{align}
	\bC_\bx^\steptt
	&\equiv \bmu_\bx^\steptt [\bmu_\bx^\steptt]^\top + \bSigma_\bx^\steptt, \\
	[\bM_{i,j}]_{k,l}
	&\equiv
	\begin{cases}
		+1, & (k,l)=(i,i) ~\mathrm{or}~ (j,j), \\
		-1, & (k,l)=(i,j) ~\mathrm{or}~ (j,i), \\
		 0, & \mathrm{otherwise}.
	\end{cases}
\end{align}
Using the Taylor approximation of (\ref{EqTaylorApproximation2}), we obtain the distribution of (\ref{EqTrialDistributionEta}) at step $t+1$ with the parameter
\begin{align}
	\label{EqUpdateMuEta}
	\mu_{\eta_{i,j}}^\steptl
	&= \logistic\left( \mu^\steptt_\lambda + \frac{1}{2} \mu_\rho^\steptt C_{\eta_{i,j}}^\steptt \right),
\end{align}
where
\begin{align}
	\label{EqUpdateMuEtaA}
	C_{\eta_{i,j}}^\steptt
	&\equiv \tr \Big[ \Big(\bA(\bmu_\bmeta^\steptt, \mu_\rho^\steptt, \mu_\kappa^\steptt)^{-1} 
	   - \bC_\bx^\steptt \Big) \bM_{i,j} \Big].
\end{align}

The update equation of $\bx$ is given as
\begin{align}
	&q^\steptl(\bx)
	\propto \exp \left\langle \ln p(\bz | \bY) \right\rangle_{q^\steptl(\bmeta) q^\steptt(\lambda, \rho, \kappa, \beta, \bPhi)} \nonumber\\
	&\propto \exp \Bigg( -\frac{1}{2} \Bigg\{ 
	\bx^\top \bA(\bmu_\bmeta^\steptl, \mu_\rho^\steptt, \mu_\kappa^\steptt) \bx \nonumber\\
	&~~ + \mu_\beta^\steptt \sum_{l=1}^L 
	\left\langle \| \bW(\bphi_l)\bx - \by_l \|_2^2 \right\rangle_{q^\steptt(\bphi_l)} \Bigg\} \Bigg).
\end{align}
It becomes a Gaussian distribution. Using the Taylor approximation (\ref{EqTaylorApproximation1}), we obtain the distribution of (\ref{EqTrialDistributionX}) at step $t+1$ with the parameters
\begin{align}
	\label{EqUpdateMuX}
	\bmu_\bx^\steptl
	&= \bSigma_\bx^\steptl \left[ \mu_\beta^\steptt \sum_{l=1}^L \by_l^\top \bW^\steptt_l \right]^\top, \\
	\label{EqUpdateSigmaX}
	\bSigma_\bx^\steptl
	&= \left[  \bA(\bmu_\bmeta^\steptl, \mu_\rho^\steptt, \mu_\kappa^\steptt) + \mu_\beta^\steptt \sum_{l=1}^L \bC'^\steptt_{\bW_l} \right]^{-1},
\end{align}
where
\begin{align}
	\bC'^\steptt_{\bW_l}
	&\equiv [\bW^\steptt_l]^\top \bW^\steptt_l
	 + \sum_{k,k'} [\bSigma_{\bphi_l}^\steptt]_{k,k'} {[\bW'^\steptt_{l,k}}]^\top \bW'^\steptt_{l,k'}.
\end{align}

The update equation of $\lambda, \rho, \kappa, \beta$ is given as
\begin{align}
	&q^\steptl(\lambda, \rho, \kappa, \beta)
	 \propto \exp \left\langle \ln p(\bz | \bY) \right\rangle_{q^\steptl(\bx, \bmeta) q^\steptt(\bPhi)} \nonumber\\
	&\propto
	\lambda^{a_\lambda^\stepOO - 1} \rho^{a_\rho^\stepOO - 1} \kappa^{a_\kappa^\stepOO - 1} \beta^{a_\beta^\stepOO + \frac{1}{2} L N_\by}
	 \exp \Bigg( \nonumber\\
	&~~ - \left\{ b_\lambda^\stepOO + \sum_{i \sim j} (1 - \mu_{\eta_{i,j}}^\steptl) \right\} \lambda \nonumber\\
	&~~ - \left\{ b_\rho^\stepOO + \frac{1}{2} \tr \bC_\bx^\steptl \bA(\bmu_\bmeta^\steptl, 1, 0) \right\} \rho  \nonumber\\
	&~~ - \left\{ b_\kappa^\stepOO + \frac{1}{2} \tr \bC_\bx^\steptl \right\} \kappa \nonumber\\
	&~~ -\Bigg\{ b_\beta^\stepOO + \frac{1}{2} \sum_{l=1}^L \Big\langle 
	 \tr \bC_\bx^\steptl \bW(\bphi_l)^\top \bW(\bphi_l) \nonumber\\
	&~~~~~~ - 2 \by_l^\top \bW(\bphi_l) \bmu_\bx^\steptl + \by_l^\top \by_l
	 \Big\rangle_{q^\steptt(\bphi_l)} \Bigg\} \beta \nonumber\\
	&~~ + \frac{1}{2} \left\langle \ln \left| \bA(\bmeta, \rho, \kappa) \right| \right\rangle_{q^\steptl(\bmeta)}
	    + N_\bmeta \ln \logistic(\lambda) \Bigg).
\end{align}
Using Taylor approximations (\ref{EqTaylorApproximation1}), (\ref{EqTaylorApproximation2}), and (\ref{EqTaylorApproximation3}), we obtain the distribution of (\ref{EqTrialDistributionLambda}) at step $t+1$ with parameters
\begin{align}
	\label{EqUpdateALambda}
	a_\lambda^\steptl
	&= a_\lambda^\stepOO + N_\bmeta \mu_\lambda^\steptt \logistic(-\mu_\lambda^\steptt), \\
	\label{EqUpdateBLambda}
	b_\lambda^\steptl
	&= b_\lambda^\stepOO + \sum_{i \sim j} (1 - \mu_{\eta_{i,j}}^\steptl), \\
	\label{EqUpdateARho}
	a_\rho^\steptl
	&= a_\rho^\stepOO
	 + \frac{\mu_\rho^\steptt}{2} \tr \bA(\bmu_\bmeta^\steptl, \mu_\rho^\steptt, \mu_\kappa^\steptt)^{-1} \bA(\bmu_\bmeta^\steptl, 1, 0)\!\! \\
	\label{EqUpdateBRho}
	b_\rho^\steptl
	&= b_\rho^\stepOO + \frac{1}{2} \tr \bC_\bx^\steptl \bA(\bmu_\bmeta^\steptl, 1, 0), \\
	\label{EqUpdateAKappa}
	a_\kappa^\steptl
	&= a_\kappa^\stepOO + \frac{\mu_\kappa^\steptt}{2}  \tr \bA(\bmu_\bmeta^\steptl, \mu_\rho^\steptt, \mu_\kappa^\steptt)^{-1} \\
	\label{EqUpdateBKappa}
	b_\kappa^\steptl
	&= b_\kappa^\stepOO + \frac{1}{2} \tr \bC_\bx^\steptl, \\
	\label{EqUpdateABeta}
	a_\beta^\steptl
	&= a_\beta^\stepOO + \frac{1}{2} L N_\by, \\
	\label{EqUpdateBBeta}
	b_\beta^\steptl
	&= b_\beta^\stepOO + \frac{1}{2} \sum_{l=1}^L 
	 \left( \tr \bC_\bx^\steptl \bC'^\steptt_{\bW_l} - 2 \by_l^\top {\bW^\steptt_l} \bmu_\bx^\steptl + \by_l^\top \by_l \right).
\end{align}

The update equation of $\bPhi$ is given as
\begin{align}
	&q^\steptl(\bPhi)
	 \propto \exp \left\langle\ln p(\bz | \bY) \right\rangle_{q^\steptl(\bx, \bmeta) q^\steptt(\lambda, \rho, \kappa, \beta)} \nonumber\\
	&\propto \exp \Bigg(
	 - \frac{1}{2} \sum_{l=1}^L \bigg\{
	[\bphi_l - \bmu_{\bphi_l}^\stepOO]^\top [\bSigma_{\bphi_l}^\stepOO]^{-1} [\bphi_l - \bmu_{\bphi_l}^\stepOO] \nonumber\\
	&~~ + \mu_\beta^\steptt \left\{ \tr \bC_\bx^\steptl \bW(\bphi_l)^\top \bW(\bphi_l) -2 \by_l^\top \bW(\bphi_l) \bmu_\bx^\steptl \right\}
	\bigg\} \Bigg).
\end{align}
Using the Taylor approximation (\ref{EqTaylorApproximation1}), we obtain the distribution of (\ref{EqTrialDistributionPhi}) at step $t+1$ with parameters
\begin{align}
	\label{EqUpdateMuPhi}
	\bmu_{\bphi_l}^\steptl
	&= \bSigma_{\bphi_l}^\steptl \left[ [\bSigma_{\bphi_l}^\stepOO]^{-1} \bmu_{\bphi_l}^\stepOO
	+ \mu_\beta^\steptt [ \bC''^\steptl_{\bphi_l} \bmu_{\bphi_l}^\steptt - \bC'^\steptl_{\bphi_l} ] \right], \\
	\label{EqUpdateSigmaPhi}
	\bSigma_{\bphi_l}^\steptl
	&= \left[ [\bSigma_{\bphi_l}^\stepOO]^{-1} + \mu_\beta^\steptt \bC''^\steptl_{\bphi_l} \right]^{-1},
\end{align}
where
\begin{align}
	[\bC'^\steptl_{\bphi_l}]_k
	&\equiv \frac{1}{2} \tr \bC_\bx^\steptl \left[ [\bW^\steptt_l]^\top \bW'^\steptt_{l,k} + [\bW'^\steptt_{l,k}]^\top \bW^\steptt_l \right] \nonumber\\
	&~~ - \by_l^\top \bW'^\steptt_{l,k} \bmu_\bx^\steptl, \\
	[\bC''^\steptl_{\bphi_l}]_{k,k'}
	&\equiv \tr \bC_\bx^\steptl [\bW'^\steptt_{l,k}]^\top \bW'^\steptt_{l,k'}.
\end{align}

\end{document}